%% file: acl_latex.tex
\pdfoutput=1

\documentclass[11pt]{article}

\usepackage[preprint]{acl}

\usepackage{times}
\usepackage{latexsym}

\usepackage[T1]{fontenc}

\usepackage[utf8]{inputenc}

\usepackage{microtype}

\usepackage{inconsolata}

\usepackage{graphicx}

\usepackage{booktabs}
\usepackage{amsmath}
\usepackage{algorithm}
\usepackage{algpseudocode}

\usepackage{multirow}
\usepackage{amssymb}
\usepackage{tabularx}
\usepackage{xcolor}
\usepackage{arydshln}
\usepackage{enumitem}
\usepackage{soul}

\definecolor{shadowred}{HTML}{F8CECC}

\title{DIVE into MoE: Diversity-Enhanced Reconstruction of Large Language Models from Dense into Mixture-of-Experts}

\author{
    Yuchen Feng$^{1,2}$\footnotemark[1], 
    Bowen Shen$^{1,2}$\thanks{$\quad$ Equal Contribution.}, 
    Naibin Gu$^{1,2}$, 
    Jiaxuan Zhao$^{1,2}$, \\
    \textbf{Peng Fu}$^{1,2}$\thanks{$\quad$ Corresponding Author.}, 
    \textbf{Zheng Lin}$^{1,2}$, 
    \textbf{Weiping Wang}$^{1}$\\
    $^1$Institute of Information Engineering, Chinese Academy of Sciences, Beijing, China\\
    $^2$School of Cyber Security, University of Chinese Academy of Sciences, Beijing, China\\
   \texttt{\{fengyuchen,shenbowen,fupeng\}@iie.ac.cn}
}

\begin{document}

\maketitle
\begin{abstract}

Large language models (LLMs) with the Mixture-of-Experts (MoE) architecture achieve high cost-efficiency by selectively activating a subset of the parameters. Despite the inference efficiency of MoE LLMs, the training of extensive experts from scratch incurs substantial overhead, whereas reconstructing a dense LLM into an MoE LLM significantly reduces the training budget. However, existing reconstruction methods often overlook the diversity among experts, leading to potential redundancy. In this paper, we come up with the observation that a specific LLM exhibits notable diversity after being pruned on different calibration datasets, based on which we present a \textbf{Div}ersity-\textbf{E}nhanced reconstruction method named DIVE. The recipe of DIVE includes domain affinity mining, pruning-based expert reconstruction, and efficient retraining. Specifically, the reconstruction includes pruning and reassembly of the feed-forward network (FFN) module. After reconstruction, we efficiently retrain the model on routers, experts and normalization modules. We implement DIVE on Llama-style LLMs with open-source training corpora. Experiments show that DIVE achieves training efficiency with minimal accuracy trade-offs, outperforming existing pruning and MoE reconstruction methods with the same number of activated parameters. Code is available at: \url{https://github.com/yuchenblah/DIVE}.

\end{abstract}

\input{body/intro}

\input{body/method}

\input{body/experiment}

\input{body/related}

\input{body/conclusion}

\input{body/after_main_paper}

\bibliography{custom}

\clearpage

\appendix

\input{body/appendix}

\end{document}

%% file: body/intro.tex
\section{Introduction}

\begin{figure}[t]
    \centering
    \includegraphics[width=1\linewidth]{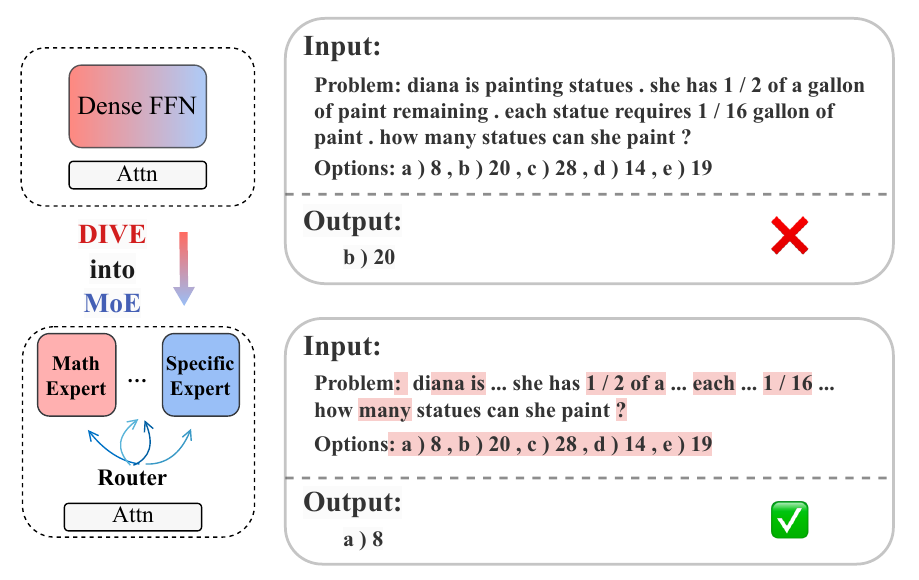}
    \caption{Example inputs and outputs of a dense LLM and DIVE, sampled from MathQA~\cite{amini-etal-2019-mathqa}. Tokens highlighted with the \sethlcolor{shadowred}\hl{red background} are routed to the Math expert. DIVE benefits from the inherent diversity in the dense LLM with specialized experts.}
\label{fig:intro}
\end{figure}

Large Language Models (LLMs) based on Transformer~\cite{c:22} have demonstrated outstanding capabilities across various domains. However, prevailing dense LLMs with all parameters activated during inference involve significanst computational and memory overheads, hindering the practical deployment of LLM-based services~\cite{zhu2023survey,gu-etal-2024-light}. While static compression like structured pruning can reduce these overheads, they often incurs downgraded capacity and generalization abilities.

Unlike reducing the size of pre-trained LLMs, Mixture-of-Experts (MoE) LLMs, such as Mixtral 8x7B~\cite{jiang2024mixtral}, enable scalable pre-training by activating only a subset of parameters during inference, achieving superior performance with the same or fewer active parameters compared to dense LLMs~\cite{lepikhin2020gshard}. MoE models typically differ from dense LLMs in the feed-forward network (FFN), replacing the single FFN with multiple smaller FFN experts. These experts are selectively activated via a parameterized router, while other components like embeddings, language modeling heads, and multi-head attention (MHA) modules remain unchanged.

Although MoE models share similarities with dense LLMs, they are usually pre-trained from scratch, which involves significant costs and potential instability~\cite{wei2024skywork}. To mitigate these issues, there has been increasing interest in converting dense models into MoE architectures. However, existing methods, like duplicating FFN modules~\cite{komatsuzaki2022sparse} and random splitting~\cite{zhu2024llama}, often produce homogeneous experts that lack diversity. This necessitates substantial retraining to enhance expert diversity and achieve optimal performance. In this paper, we aim to tackle this challenge from the perspective of expert reconstruction, addressing the following key question:

\textit{How can we reconstruct a dense LLM into MoE architecture while enhancing expert diversity and significantly reducing training costs?}

In pursuit of enhancing diversity, we investigate structured pruning approaches, which derive smaller but potentially less generalizable models from existing LLMs. Pruning typically relies on a calibration set to assess weight importance~\cite{an2024fluctuation}. Given the aggressive nature of pruning and the limited size of calibration sets compared to pre-training corpora, certain pruning methods exhibit sensitivity to the calibration set. We observe that LLMs pruned with different calibration sets exhibit notable diversity across evaluation domains, and models pruned on specific domain data struggling to generalize. While such sensitivity can be a drawback, it effectively uncovers the inherent diversity in the weights of dense LLMs.

Building on the above insights from pruning, we propose a \textbf{Div}ersity-\textbf{E}nhancing expert reconstruction method of LLM named DIVE.  As illustrated in Figure~\ref{fig:intro}, DIVE leverages the inherent diversity of the original LLMs, and construct experts with varying capabilities. We firstly use domain affinity mining to identify expert domains and deploys pruning-based expert reconstruction. Then, we design an efficient retraining pipeline to recover the model's capabilities, only the reconstructed modules (routers, experts, and normalizations) need to be retrained with parameter-efficient tuning (PEFT). As a summary, our contributions are as follows:

\begin{itemize}
    \item We introduce DIVE, an effective MoE reconstruction method for dense LLMs that enhances expert diversity, inspired by static pruning techniques. 
    \item We propose an efficient retraining method, designed to minimize training costs. Only less than $1$\% of the parameters are required to be tuned. 
    \item Experiments show the effectiveness and robustness of DIVE. With a 50\% sparsity of the original FFN, DIVE outperforms existing pruning and MoE reconstruction methods, excelling in both language modeling and various downstream tasks.
\end{itemize}

%% file: body/method.tex
\begin{figure*}[t]
    \centering
    \includegraphics[width=1\linewidth]{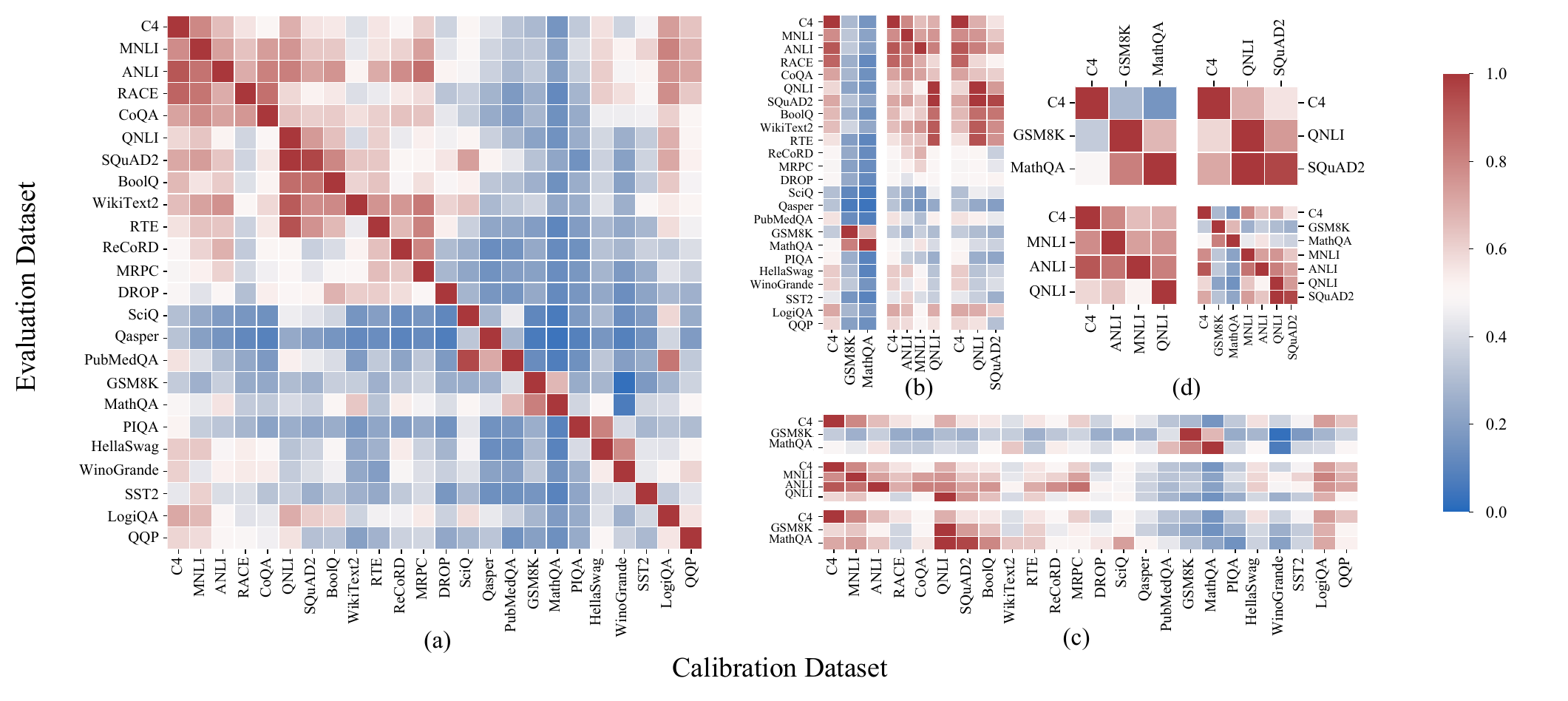}
    \caption{The heatmap shows normalized perplexity (PPL) with a 75\% FFN pruning ratio on LLaMA2-7B. According to the definition of normalized PPL in Eq.~\ref{eq:nor_ppl}, the redder the color, the better the performance. Subfigure (a) presents the evaluation results across different calibration sets. In (b), we highlight calibration sets from (a) that lead to strong performance on other evaluation tasks, and (c) correspondingly shows that certain evaluation tasks benefit significantly from specific calibration sets. (d) isolates these tasks for a clearer view of the emerging patterns.}
    \label{fig:heatmap}
\end{figure*}

\section{Preliminary}
\label{sec:preliminary}

\subsection{Dense FFN in Transformer}
\label{sec:pre-dense}

The FFNs in standard Transformer-based dense LLMs~\cite{c:22} consist of fully connected layers that operate independently for each token. Given an input vector $x$, the forward pass through the FFN is:
\begin{equation}
    FFN(x) = F_\text{down}\left(\sigma\left(F_\text{up}\left(x\right)\right)\right),
\end{equation}
where $F_\text{up}$ and $F_\text{down}$ are linear transformations, and $\sigma$ stands for activation functions (e.g., ReLU~\cite{agarap2018deep}). For the SwiGLU activation used in Llama-style LLMs,
\begin{equation}
    \sigma(F_\text{up}(x))=\varsigma(F_\text{gate}) \odot F_\text{up}(x),
\end{equation}
where $F_\text{gate}$ is an additional linear transformation with the same shape as $F_\text{up}$, $\varsigma$ is the Swish function and $\odot$ denotes element-wise multiplication~\cite{shazeer2020glu}.

\subsection{Sparse FFN in Mixture-of-Experts}
\label{sec:pre-moe}

Different from dense LLMs, in Transformer-based MoE LLMs, the FFN layers are typically replaced with MoE layers consisting of a router $R$ and $n$ original FFN-type experts \{$FFN_1, FFN_2, \ldots, FFN_n$\}, with only the top-$k$ being activated. The router, often a lightweight feed-forward neural network, learns to optimally allocate inputs to experts based on different data distributions.

Given an input $x$, the output of an MoE layer is the weighted sum of the outputs from selected experts:
\begin{equation}
    {o(x)} = \sum_{i=1}^{k} w_i(x) \cdot FFN_i(x),
\end{equation}
where $w_i(x) = \mathrm{Softmax}(\mathrm{TopK}(z(x)))_i$ representing the normalized weight for each expert with the routing logits $z(x)$ produced by the router.

\section{Method}

\begin{figure*}[t]
    \centering
    \includegraphics[width=1\linewidth]{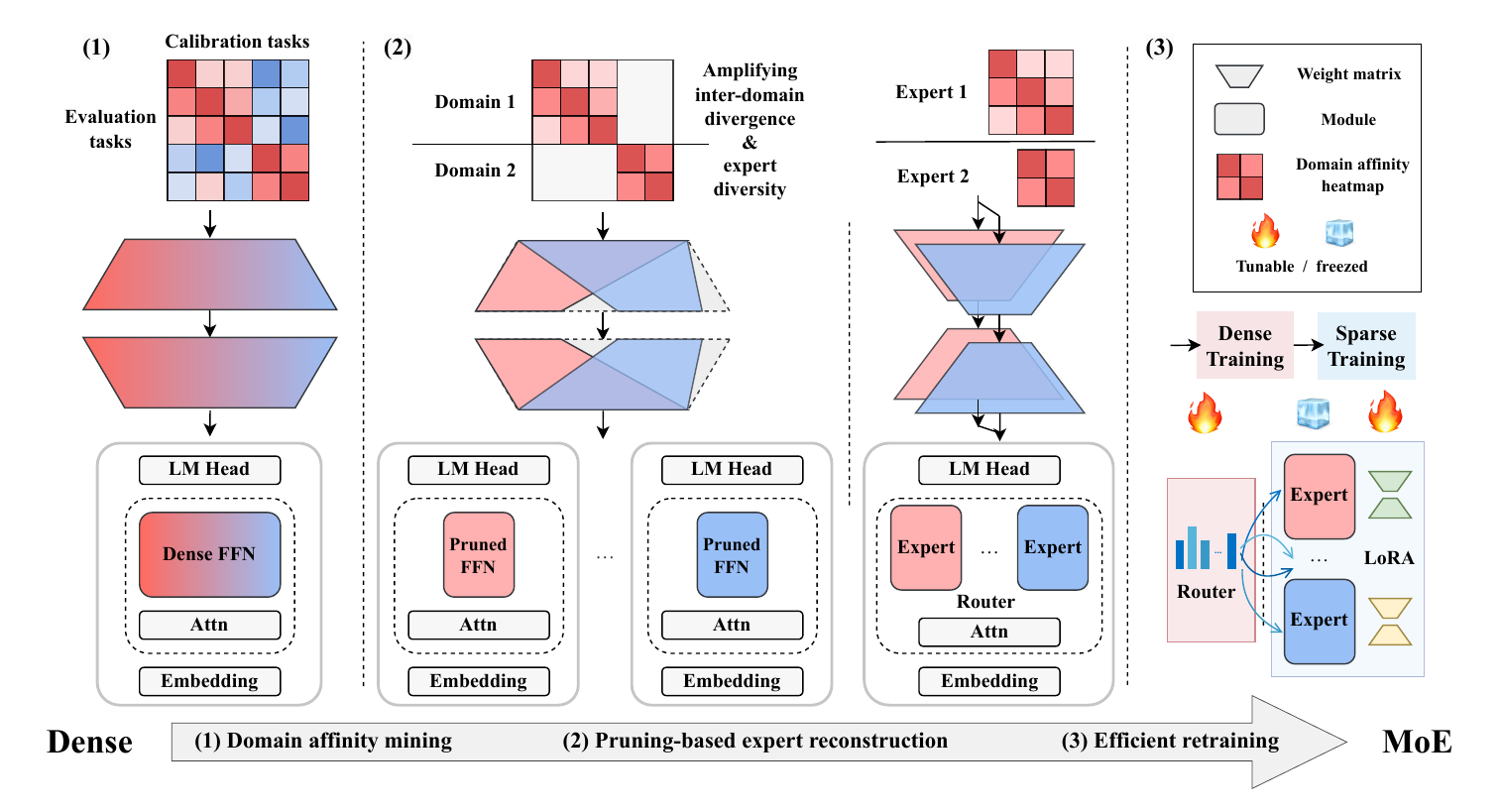}
    \caption{Demonstration of diversity-enhanced reconstruction of LLMs from dense to MoE. In each lane, the upper part exhibits the expert division with the domain affinity heatmap. The reconstruction of FFNs is concretely illustrated in the middle part. And, the modifications to the overall LLM architecture are depicted in the lower part.}
    \label{fig:overview}
\end{figure*}

Aiming at improving expert diversity in MoE reconstruction, we introduce DIVE. Our method consists of three components: domain affinity mining, pruning-based expert reconstruction, and efficient retraining.

\subsection{Domain Affinity Mining}
\label{sec:domain-affinity}

We begin with aligning the expert reconstruction with structured pruning for LLMs, applying the same pruning ratio to each FFN. An effective pruning method is FLAP~\cite{an2024fluctuation}, which measures channel importance by assessing the fluctuation across the calibration dataset in the hidden state features. From the observations of performance gaps in FLAP across different calibration datasets (Appendix~\ref{sec:appendix-a-2}), we hypothesize that \textit{fluctuation is sensitive to the domain of the calibration sets, resulting in diversity of pruning outcomes}. To verify this, we conduct domain affinity mining on LLaMA2-7B using 24 datasets from various domains (Appendix~\ref{sec:appendix-d-1}).

For each calibration task $t_i$ in the set of tasks $\mathcal{T}$, given $N$ samples, we calculate the pruning mask $\mathbf{M}^\ell_{t_i}$ base on the fluctuation variance. The importance score for each weight group $\mathbf{W}^\ell_{:,j}$ is computed as follows:
\begin{equation}
    \mathbf{S}^{\ell}_{:,j,} = \frac{1}{N-1} \sum_{n=1}^N (\mathbf{X}^{\ell}_{n,j,:,t_i} - \overline{\mathbf{X}}^{\ell}_{:,j,:,t_i})^2  \cdot ||\mathbf{W}^{\ell}_{:,j}||^2_2,
    \label{eq:fluc}
\end{equation}
where $\mathbf{X}^{\ell}_{n,j,:,t_i}$ and $\overline{\mathbf{X}}^{\ell}_{:,j,:,t_i}$ represent the hidden state feature for the $n$-th sample and the baseline fluctuation in the $j$-th channel. After pruning, channels with the highest scores are retrained. Importantly, our pruning targets the intermediate dimensions of each FFN, focusing on reconstructing the MoE layer, as discussed in Section~\ref{sec:pre-moe}.

To evaluate the pruning results, we measure the normalized perplexity (PPL) on evaluation tasks for models pruned using various calibration task sets. Given the variability of PPL values across different evaluation tasks, we introduce the definition of normalized PPL as follows:
\begin{equation}
    \mathrm{norm}(p)_{i,j} = \frac{\mathrm{min}(p)_{:,j}}{p_{i,j}},
    \label{eq:nor_ppl}
\end{equation}
where $p_{i,j}$ denotes the original PPL of the $i$-th calibration datasets on the $j$-th evaluation datasets, and $\mathrm{min}(p)_{:,j}$ denotes the minimum PPL on the $j$-th evaluation dataset.

Figure~\ref{fig:heatmap} shows evaluation results after pruning across all calibration sets with the FFN pruning ratio of 75\%. We identify key phenomena that shed light on the effects of pruning across different calibration sets:

\paragraph{(1) Calibration sets from similar domains exhibit strong correlations.}
For mathematic QA tasks (MathQA and GSM8K), the model pruned with one calibration set demonstrates strong generalization on the other task, indicating that pruning can effectively capture domain-specific expertise.

\paragraph{(2) Affinity is more sensitive to data than task types.}
The affinity between QNLI and other NLI tasks (e.g., ANLI and MNLI) is not as strong as that with SQuAD2, which stems from both QNLI and SQuAD2 being derived from SQuAD1.1 data \cite{rajpurkar-etal-2018-know}.

\paragraph{(3) C4 exhibits generalization as a calibration set.}
C4, a cleaned web crawl corpus \cite{raffel2020exploring}, shows broad generalization when used as a calibration set, reflecting its diverse linguistic features and robustness.

We further perform domain affinity mining on various LLM backbones (Appendix~\ref{sec:appendix-d-2}). These findings back up our hypothesis and motivate pruning with affinity domains to achieve diversity-enhanced MoE reconstruction.

\subsection{Pruning-Based Expert Reconstruction}

In this section, we describe how to conduct pruning-based reconstruction of domain-specific experts upon the above domain affinity mining. The reconstruction process is defined as follows: Given a dense LLM $\mathcal{M}$, obtain a converted sparse MoE LLM $\tilde{\mathcal{M}}$, with FFN modules replaced by MoE layers. Each MoE layer is composed of a router and multiple experts, which are pruned from the original dense FFNs.

We begin by calculating the domain correlations. The correlation between each pair of calibration sets is computed with the Pearson correlation coefficient:
\begin{equation}
    \mathrm{corr}(d^1, d^2) = \frac{\text{cov}(d^1, d^2)}{\sigma(d^1) \sigma(d^2)},
    \label{eq:corr}
\end{equation}
where $\text{cov}(d^1, d^2)$ is the covariance between $d^1$ and $d^2$, and $\sigma(d^1)$ and $\sigma(d^2)$ are the standard deviations of $d^1$ and $d^2$, respectively. The correlation value measures the linear correlation between two datasets, with values ranging from -1 to 1.

We use hierarchical clustering to group the calibration datasets into 8 clusters based on Eq.~\ref{eq:corr}, and construct final calibration datasets for pruning domain-specific experts by uniformly mixing data within each cluster. For each cluster, we prune an LLM on intermediate dimensions of the FFNs on it. The pruned FFNs are then restructured as experts, with each layer guided by a randomly initialized noisy router to form the MoE layers, replacing the original FFNs.

\subsection{Efficient Retraining}
\label{sec:efficient-retraining}

Since the LLM is partially reorganized, it is essential to retrain the model on a large-scale dataset to recover its general performance. Considering the sparse gradients and non-differentiability raised by the top-$k$ operation in routers, we perform a two-stage retraining: \textbf{(1) dense training} for routers using a tiny amount of data, with all experts being activated, and \textbf{(2) sparse training} for experts and normalization modules on a large dataset with PEFT, only activating the correlated experts.

During the dense training stage, we introduce a temperature coefficient $t$ to routers as in Eq.\ref{eq:temp}. This operation aims to align the routing behavior better with sparse inference. Further clarifications of the two-stage retraining design are in Appendix~\ref{sec:appendix-b-2} and ~\ref{sec:appendix-b-3}.

\begin{equation}
    w_i(x) = \mathrm{Softmax}\left(\frac{z(x)}{t}\right)_i.
    \label{eq:temp}
\end{equation}

We use Low-Rank Adaptation (LoRA)~\cite{hu2022lora} to retrain the experts, with fully fine-tuning the routers and the normalization modules. There is no need to update the parameters of MHA and other modules, as verified by experiments in Section~\ref{sec:retraining}.

\subsection{Overview of DIVE}

Generally, DIVE is designed with the workflow described in Figure~\ref{fig:overview} and Algorithm \ref{alg:dive}. With the original dense LLM $\mathcal{M}$ and target expert number $N$, $\mathcal{M}$ is first pruned and evaluated on different pairs of tasks. Then, the clustering-based data selection is performed with the distance metric of task correlation. With the grouped datasets $\mathcal{G}$, $\mathcal{M}$ is pruned with calibration data sampled from $\mathcal{G}$ into $N$ dense LLM $\{{\mathcal{M}^\prime}_1, \cdots, {\mathcal{M}^\prime}_N\}$. As ${\mathcal{M}^\prime}_N$ are with the same parameters with the only exception of FFN modules, they can be merged into an MoE architecture $\tilde{\mathcal{M}}$. Finally, a two-stage retraining on $\tilde{\mathcal{M}}$ FFN modules with PEFT recovers the performance and yields the final MoE LLM $\tilde{\mathcal{M}^{*}}$.

\begin{algorithm}
    \caption{Workflow of DIVE}
    \begin{algorithmic}[1] 
    \Require dense LLM $\mathcal{M}$, expert number $N$, tasks $\mathcal{T}$
    \Ensure MoE LLM $\tilde{\mathcal{M}^{*}}$
        \For {$(t_i, t_j) \in \mathcal{T} \times \mathcal{T}$}
        \State $P_{i,j} \gets \textsc{PPL}(\textsc{Prune}(\mathcal{M}, t_i),t_j)$
        \EndFor
        \State $\mathcal{G} \gets \textsc{Cluster}(\textsc{norm}(P), N, \text{dist}=corr)$
        \For {$i \gets 1 \ \text{to} \ N$}
        \State $\mathcal{C}_i \gets \textsc{Sample}(\mathcal{G}_i)$
        \State ${\mathcal{M}^\prime}_i \gets \textsc{Prune}(\mathcal{M}, \mathcal{C}_i)$
        \EndFor
        \State $\mathcal{M}^\prime \gets \textsc{Reconstruction}(\{{\mathcal{M}^\prime}_1, \cdots, {\mathcal{M}^\prime}_N\})$
        \State $\textsc{Freeze}(\mathcal{M}^\prime) \ \text{and} \ \textsc{Unfreeze}(\mathcal{M}^\prime.Router)$
        \State $\tilde{\mathcal{M}} \gets \textsc{Train}(\mathcal{M}^\prime.Router)$
        \State $\textsc{Freeze}(\tilde{\mathcal{M}}) \ \text{and} \ \textsc{Unfreeze}(\tilde{\mathcal{M}}.LN)$
        \State $\tilde{\mathcal{M}^{*}} \gets \textsc{Train}(\tilde{\mathcal{M}}.LN \cup \tilde{\mathcal{M}}.FFN.LoRA)$
        \State\Return $\tilde{\mathcal{M}^{*}}$
    \end{algorithmic}
    \label{alg:dive}
\end{algorithm}

%% file: body/experiment.tex
\input{tables/ppl_results}

\section{Experiments}
\subsection{Experimental Settings}

\paragraph{Baselines.}
We implement experiments on TinyLlama-1.1B~\cite{zhang2024tinyllama}, a Llama-style LLM with dense SwiGLU FFNs~\cite{shazeer2020glu}. We compare DIVE with existing structured pruning and MoE reconstruction methods for LLMs, including LLM-Pruner~\cite{ma2023llm}, FLAP~\cite{an2024fluctuation}, and LLaMA-MoE~\cite{zhu2024llama}. We set the same size of 50\% of a single original FFN for all methods and apply identical retraining procedures. FLAP is implemented on a mixed calibration set of all 24 datasets. Additionally, to make LLaMA-MoE adaptable to reconstruct MoE models with various sizes, we make adjustments to its expert construction method, as detailed in Appendix~\ref{sec:appendix-a}.

\paragraph{Retraining Setup.}
We use the SlimPajama dataset for retraining, which is a cleaned and deduplicated version of the RedPajama dataset, containing 627B tokens of training data~\cite{cerebras2023slimpajama}. Experiments of MoE reconstruction methods are conducted on randomly sampled 0.5B tokens for dense training of routers and 5B tokens for sparse training, with a sequence length of 1024, if not specified. Experiments of pruning methods are aligned with the sparse training stage. We set the temperature coefficient to 0.05 for DIVE 1/8 and 0.5 for DIVE 2/8. Detailed hyper-parameters are provided in Appendix~\ref{sec:appendix-b-1}.

\input{tables/few_shot}

\paragraph{Evaluation Metrics.}
The language modeling ability is evaluated with perplexity on WikiText2~\cite{merity2016pointer} and LAMBADA~\cite{paperno2016lambada}. We use lm-evaluation-harness~\cite{eval-harness} to evaluate the performance on downstream tasks. Following LLaMA-MoE~\cite{zhu2024llama} and FLAP~\cite{an2024fluctuation}, our evaluation covers $11$ commonly used benchmarks for LLMs: 0-shot of SciQ~\cite{SciQ}, PIQA~\cite{bisk2020piqa}, WinoGrande~\cite{WINOGRANDE}, ARC-e ~\cite{clark2018think}, MathQA~\cite{amini-etal-2019-mathqa}, LogiQA \citep{liu2020logiqa}, BoolQ~\cite{clark-etal-2019-boolq} and OBQA~\cite{OpenBookQA2018}, 25-shot ARC-c \citep{clark2018think}, 10-shot HellaSwag~\cite{zellers2019hellaswag} and 5-shot MMLU~\cite{hendryckstest2021}.

\subsection{Main Results}
\subsubsection{Language Modeling}

Table~\ref{tab:ppl-results} presents the language modeling results of DIVE and other methods after retraining. Models are grouped by single FFN size: 50\% $\times$ 1 and 25\% $\times$ 2. The perplexity scores demonstrate that DIVE 1/8 significantly outperforms other methods with the same number of activated parameters. In particular, compared to the second-best results, DIVE 1/8 reduces the perplexity by 8.38 on LAMBADA, and by 0.99 and 0.89 on WikiText2 with sequence lengths of 1024 and 2048, respectively.

\begin{figure}[htbp]
    \centering
    \includegraphics[width=1\linewidth]{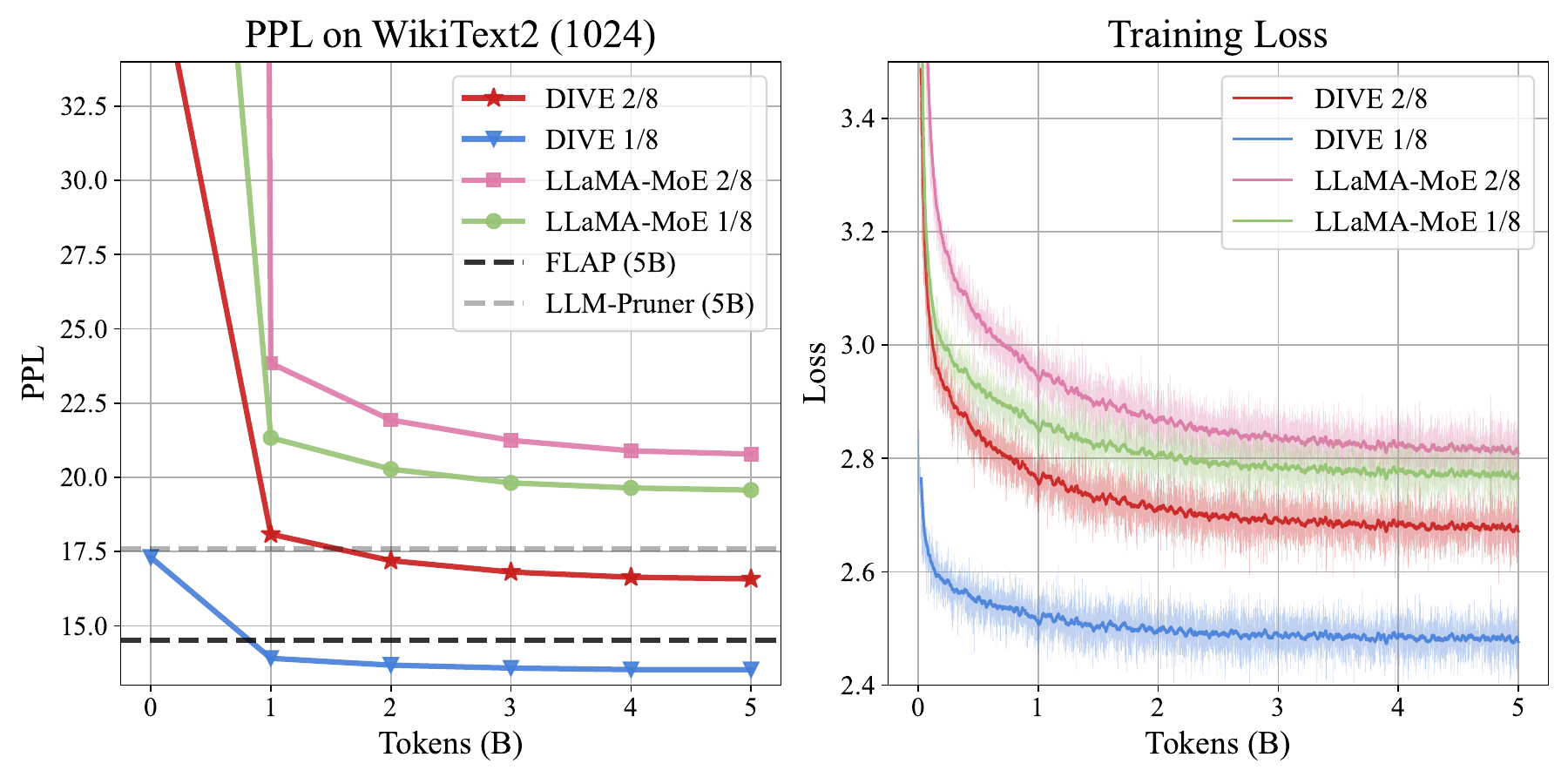}
    \caption{Perplexity curves of DIVE and the baselines, consisting of perplexity on WikiText2 (left) and training loss for DIVE and LLaMA-MoE (right). Loss curves are smoothed with a 0.05B token window.}
\label{fig:ppl_curve}
\end{figure}

We further analyze the performance of MoE reconstruction methods during retraining, focusing on the perplexity for WikiText2 with the sequence length of 1024 and the training loss. As shown in Figure~\ref{fig:ppl_curve}, our models consistently outperform LLaMA-MoE counterparts, during retraining with 1B to 5B tokens. Notably, DIVE 1/8 trained on merely 1B tokens surpasses FLAP trained on 5B tokens, highlighting the efficiency of our method.

\begin{figure*}[htbp]
    \centering
    \includegraphics[width=0.95\linewidth]{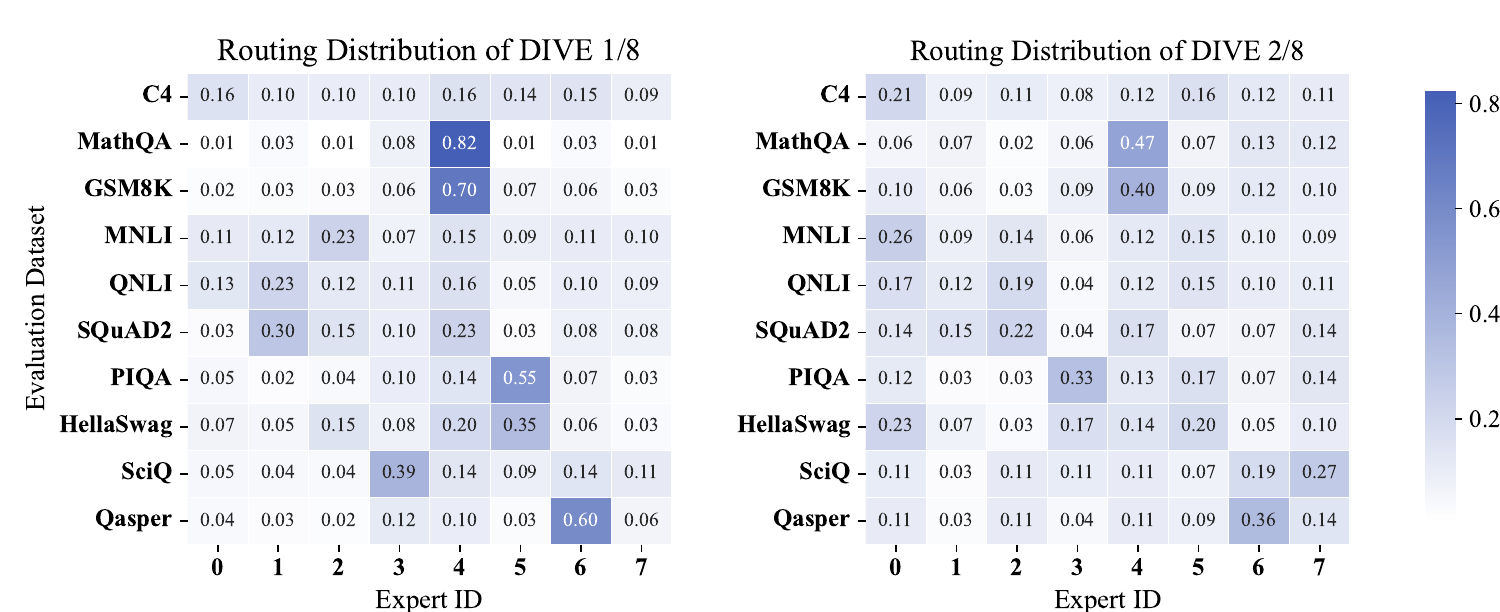}
    \caption{Routing distribution of DIVE 1/8 (left) and DIVE 2/8 (right), shown as activation ratios of the corresponding experts across all layers. The indexes of experts correspond to the domain clustering results in Appendix~\ref{sec:appendix-d-2}. Note that, since DIVE achieves expert initialization through pruning, the IDs of domain-specific experts remain consistent across layers.}
\label{fig:routing}
\end{figure*}

\begin{figure}[htbp]
    \centering
    \includegraphics[width=1\linewidth]{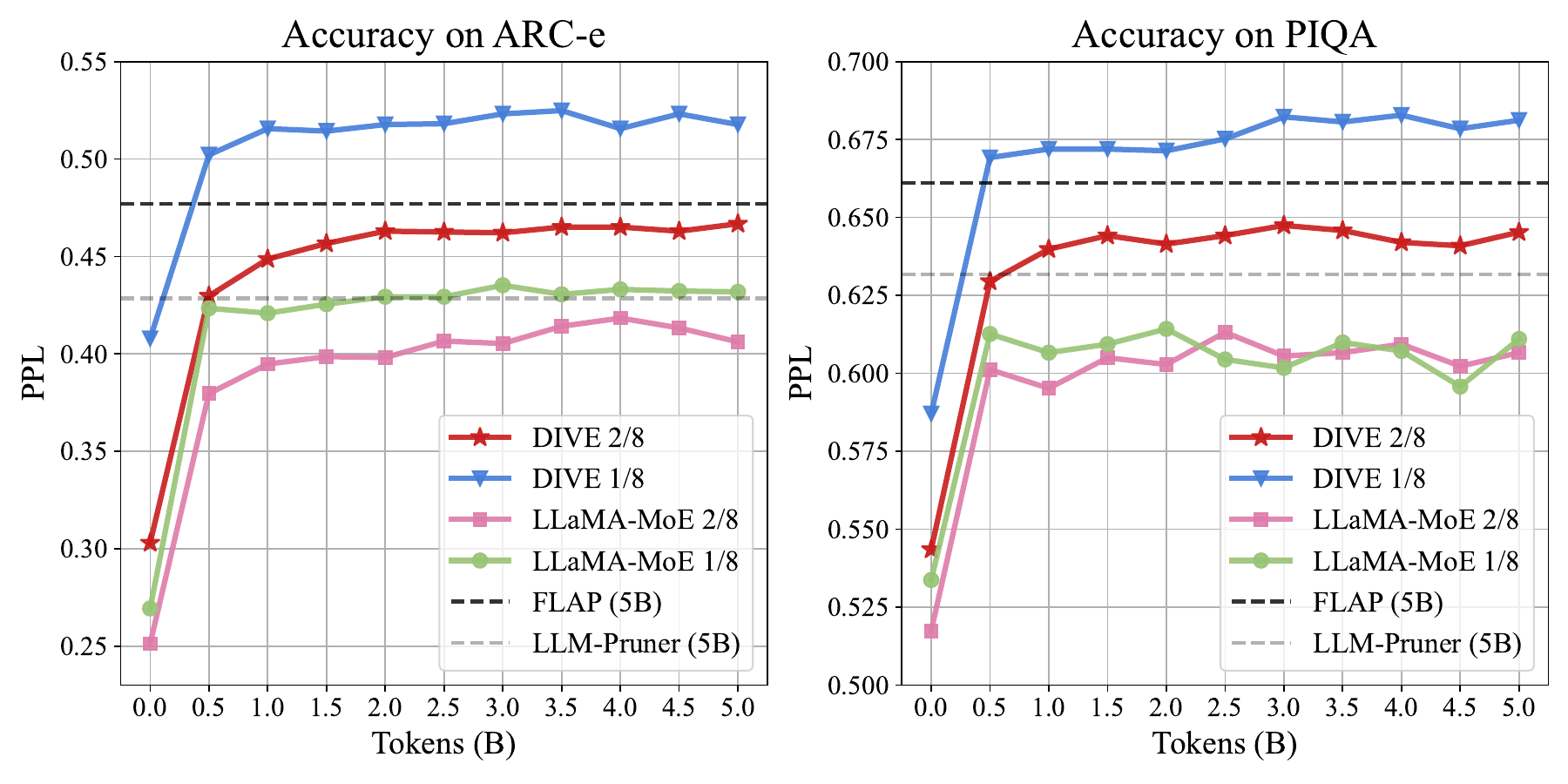}
    \caption{Performance curves of DIVE and baselines on 0-shot ARC-e (left) and PIQA (right). Solid lines indicate the performance changes of models during retraining to 5B tokens, and dashed lines represent the performance of models retrained on 5B tokens.}
\label{fig:fewshot_curve}
\end{figure}

\subsubsection{Downstream Tasks}

Table~\ref{tab:few-shot} presents the few-shot performance of accuracy across 11 downstream tasks. DIVE 1/8 achieves the best average performance among 50\% $\times$ 1 models, excelling in SciQ, PIQA, ARC-e, ARC-c, HellaSwag, OBQA, and MMLU. It improves accuracy on BoolQ by 3.55\% compared to the original model. And while DIVE 2/8 slightly lags behind DIVE 1/8 and FLAP on average, it outperforms LLM-Pruner and LLaMA-MoE 2/8 by 1.35\% and 1.68\%, despite LLM-Pruner benefiting from a more refined pruning ratio.

To illustrate the changes of model capabilities during retraining process, we present performance curves on ARC-e and PIQA in Figure~\ref{fig:fewshot_curve}. As the results on these benchmarks gradually grow, DIVE 1/8 consistently outperforms all other methods retrained on 5B tokens. Meanwhile, compared with LLaMA-MoE models, both DIVE 1/8 and DIVE 2/8 demonstrate superior performance. We further retrain the MoE models on 15B tokens to validate the effectiveness of our approach (Appendix~\ref{sec:appendix-c}).

\subsection{Diversity Analysis}
\label{sec:exp-diversity}

To assess the diversity of DIVE models, we analyze routing distributions across 24 pruning datasets, with representative results shown in Figure~\ref{fig:routing}. The values in heatmaps represent activation ratios, calculated as the frequency of tokens routed to a specific expert index across all layers, normalized by the total token activations in each evaluation set. Notably, since our MoE layers are derived through pruning, the expert indices in each layer are fully aligned with the domain clusters.

Based on the analysis, we find several observations aligned with the phenomena discussed in Section~\ref{sec:domain-affinity}:

\begin{enumerate}[itemsep=0ex, partopsep=0ex]
    \item After dense training of routers, most task tokens (such as MathQA, GSM8K, PIQA, and HellaSwag) are correctly routed to their corresponding experts.
    \item Certain tasks (QNLI and SQuAD2) exhibit sensitivity to data structure, activating Expert 1 due to shared features, while Expert 2 is mainly triggered by other NLI tasks.
    \item Tokens in C4 demonstrate strong generality, activating experts in a balanced manner with nearly uniform distribution.
    \item Compared to DIVE 1/8, DIVE 2/8 shows smoother activation ratios across experts due to the increased number of activated experts, aligning with expectations.
\end{enumerate}

These findings confirm the success of MoE diversity and routing allocation. Full heatmaps along with a case study are provided in Appendix~\ref{sec:appendix-e}.

\subsection{Ablation Studies}

\subsubsection{Domain Affinity Mining}
\label{sec:exp-domain}

\input{tables/affinity}

\begin{figure}[htbp]
    \centering
    \includegraphics[width=1\linewidth]{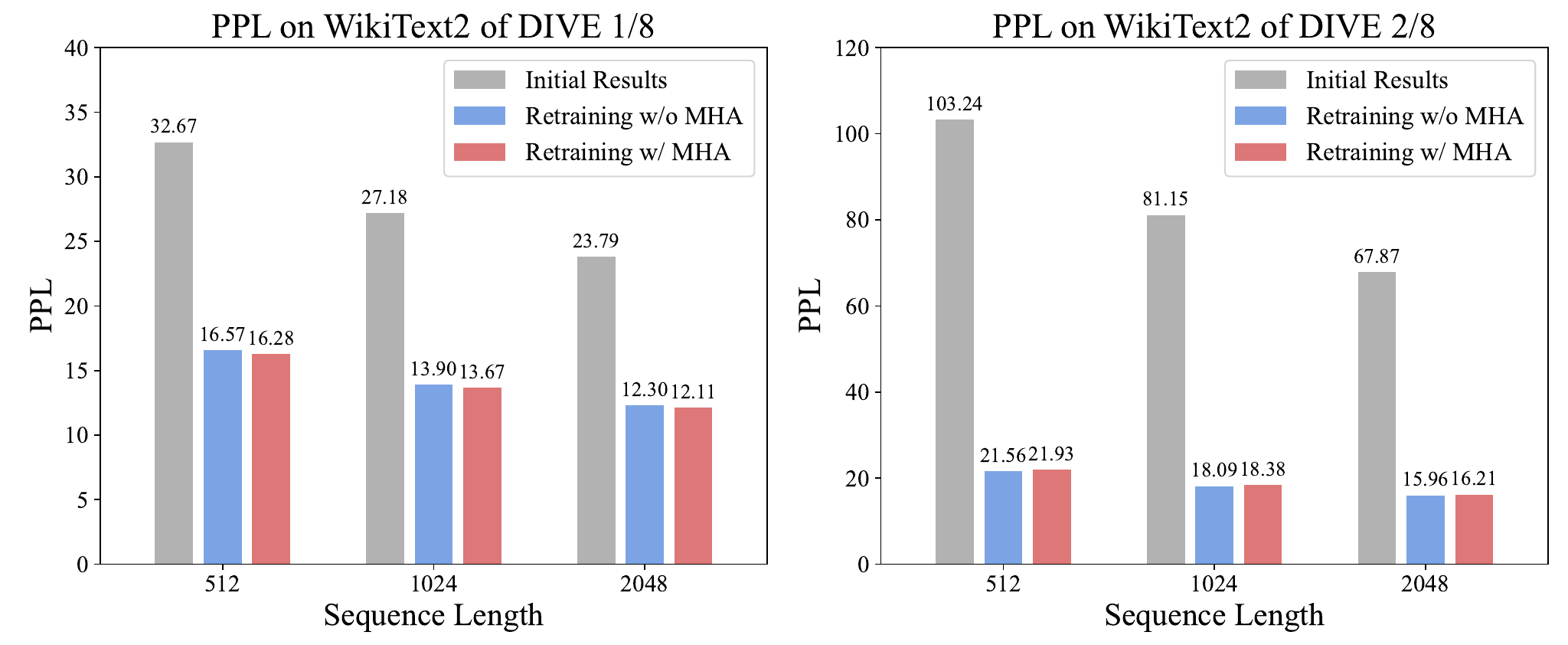}
    \caption{Comparison of DIVE 1/8 (left) and DIVE 2/8 (right) perplexity results on WikiText2, retrained on 1B tokens with and without MHA modules.}
\label{fig:retraining}
\end{figure}

To validate the effectiveness of our domain affinity mining (D.A.M) approach, which enhances the diversity of MoE models, we compare DIVE reconstructed with D.A.M against models using randomly selected calibration data from all 24 datasets. We categorize the downstream tasks into in-distribution (ID) tasks and out-of-distribution (OOD) tasks, based on whether they belong to the 24 datasets used for pruning.

As the loss and performance remain relatively stable during training from 1B to 5B tokens in Figure~\ref{fig:ppl_curve} and ~\ref{fig:fewshot_curve}, we conduct our ablation studies with 1B tokens to balance the computational efficiency and the validity of results. As shown in Table~\ref{tab:affinity}, DIVE 2/8 with D.A.M reduces the perplexity on WikiText2 with the sequence length of 1024 and LAMBADA by 1.93 and 12.58, and achieves accuracy improvements of 0.80\% and 1.52\% over the model without D.A.M. This underscores the critical role of domain affinity mining in the reconstruction process and highlights the importance of diversity in improving model performance. Detailed information of task categories and full results are provided in Appendix~\ref{sec:appendix-f}.

\subsubsection{Retraining Targets}
\label{sec:retraining}

In this section, we explain the rationale behind our efficient retraining targets. We focus on perplexity results on WikiText2 across various sequence lengths to evaluate general performance.

Figure~\ref{fig:retraining} shows the performance of the initial reconstructed MoE models and models retrained on 1B tokens with and without MHA modules. For DIVE 1/8, cutting the MHAs out of the tunable parameters only leads to less than a 0.29 degradation in performance. For DIVE 2/8, the model retrained without MHA surprisingly outperforms the one with MHA by up to 0.37. These results support the retraining targets of our method, indicating the MHA modules are unnecessary to be retrained.

%% file: tables/ppl_results.tex
\begin{table}[ht]
    \centering
    \resizebox{\columnwidth}{!}{%
    \begin{tabular}{lcccc}
    \toprule 
        \multirow{2}{*}{\textbf{Method}} & \multirow{2}{*}{\textbf{FFN Size}} & \multicolumn{2}{c}{\textbf{WikiText2}} & \multirow{2}{*}{\textbf{LAMBADA}} \\
        &  & \textbf{1024} & \textbf{2048} &  \\
        \midrule
        LLM-Pruner & 50\% & 17.59 & 15.62 & 56.66 \\
        FLAP & 50\% & 14.51 & 12.84 & 33.22 \\
        LLaMA-MoE 1/8 & 50\% $\times$ 1 & 19.57 & 17.26 & 87.27 \\
        \textbf{DIVE 1/8 (Ours)} & 50\% $\times$ 1 & \textbf{13.52} & \textbf{11.95} & \textbf{24.84} \\
        \hdashline
        LLaMA-MoE 2/8 & 25\% $\times$ 2 & 20.78 & 18.46 & 121.00 \\
        \textbf{DIVE 2/8 (Ours)} & 25\% $\times$ 2 & \textbf{16.58} & \textbf{14.67} & \textbf{51.95} \\
    \bottomrule
    \end{tabular}%
    }
    \caption{Perplexity on WikiText2 and LAMBADA language modeling. WikiText2 commonly exceeds the maximum length of the model, and the length of LAMBADA is below 1024. Best results are in \textbf{bold}.}
    \label{tab:ppl-results}
\end{table}

%% file: tables/few_shot.tex
\begin{table*}[t]
    \centering
    \resizebox{\textwidth}{!}{%
    \begin{tabular}{lcccccccc}
        \toprule
        \textbf{Method} & \textbf{FFN Size} & \textbf{SciQ} & \textbf{PIQA} & \textbf{WinoGrande} & \textbf{ARC-e} & \textbf{ARC-c (25)} & \textbf{MathQA} \\
        \midrule
        TinyLlama-1.1B & 100\% & 89.30 & 72.60 & 59.43 & 61.66 & 35.67 & 24.32 \\
        \hdashline
        LLM-Pruner & 50\% & 79.20 & 63.17 & 51.46 & 42.85 & 22.87 & 21.68\\
        FLAP & 50\% & 80.50 & 66.10 & \textbf{55.72} & 47.69 & 24.49 & \textbf{22.61} \\
        LLaMA-MoE 1/8 & 50\% $\times$ 1 & 76.30 & 61.10 & 52.17 & 43.18 & 22.70 & 21.84 \\
        \textbf{DIVE 1/8 (Ours)} & 50\% $\times$ 1 & \textbf{83.00} & \textbf{68.12} & 52.72 & \textbf{51.77} & \textbf{24.91} & 22.45 \\
        \hdashline
        LLaMA-MoE 2/8 & 25\% $\times$ 2 & 77.10 & 60.66 & \textbf{54.93} & 40.70 & 21.84 & 22.18 \\
        \textbf{DIVE 2/8 (Ours)} & 25\% $\times$ 2 & \textbf{81.10} & \textbf{64.53} & 52.49 & \textbf{46.68} & \textbf{22.70} & \textbf{23.02} \\
        \midrule
        \textbf{Method} & \textbf{FFN Size} & \textbf{HellaSwag (10)} & \textbf{LogiQA} & \textbf{BoolQ} & \textbf{OBQA} & \textbf{MMLU (5)} & \textbf{Average} \\
        \midrule
        TinyLlama-1.1B & 100\% & 46.45 & 21.51 & 55.99 & 25.20 & 26.79 & 47.18 \\
        \hdashline
        LLM-Pruner & 50\% & 33.17 & \textbf{21.04} & 59.02 & 16.20 & 24.61 & 39.57 \\
        FLAP & 50\% & 35.37 & 20.28 & \textbf{61.86} & 17.60 & 23.39 & 41.42 \\
        LLaMA-MoE 1/8 & 50\% $\times$ 1 & 33.24 & 18.74 & 61.71 & 16.60 & 25.16 & 39.34 \\
        \textbf{DIVE 1/8 (Ours)} & 50\% $\times$ 1 & \textbf{37.09} & 19.35 & 58.81 & \textbf{20.20} & \textbf{25.40} & \textbf{42.17} \\
        \hdashline
        LLaMA-MoE 2/8 & 25\% $\times$ 2 & 31.93 & 19.20 & \textbf{61.93} & 16.20 & 25.00 & 39.24 \\
        \textbf{DIVE 2/8 (Ours)} & 25\% $\times$ 2 & \textbf{32.86} & \textbf{22.12} & 59.88 & \textbf{18.00} & \textbf{26.78} & \textbf{40.92} \\
        \bottomrule
    \end{tabular}
    }
    \caption{Few-shot performance of pruning and MoE reconstruction methods on TinyLlama-1.1B, measured by accuracy on downstream tasks. The FFN size with a multiplier of $\times 1$ or $\times 2$ indicates the number of experts, while dense models do not use the multiplier. The best results are highlighted in \textbf{bold}.}
    \label{tab:few-shot}
\end{table*}

%% file: tables/affinity.tex
\begin{table}[htbp]
    \centering
    \resizebox{0.8\columnwidth}{!}{%
    \begin{tabular}{lcccc}
        \toprule
                               & \textbf{DIVE 2/8} & \textbf{- D.A.M.} \\
        \midrule
        \textbf{WikiText2 $\downarrow$} & 18.09    & 20.02 (+1.93)\\
        \textbf{LAMBADA $\downarrow$}   & 63.45    & 76.03 (+12.58)\\
        \hdashline
        \textbf{ID tasks $\uparrow$}    & 47.88    & 47.08 (-0.80) \\
        \textbf{OOD tasks $\uparrow$}   & 27.67    & 26.15 (-1.52) \\
        \bottomrule
    \end{tabular}%
    }
    \caption{Performance of DIVE 2/8 with and without D.A.M, retrained on 1B tokens. Tasks marked with downward arrows (WikiText2 and LAMBADA) are measured by perplexity and inversely correlated with performance, others (ID and OOD tasks) are downstream tasks measured by accuracy.}
    \label{tab:affinity}
\end{table}

%% file: body/related.tex
\section{Related Work}

\subsection{Dense Compression}

Statically pruning LLMs to an optimal size enables efficient deployment~\cite{xia2022structured,shen2022cost}, while posing challenges due to performance impacts and the need for retraining adjustments. Gradient-based methods have been extensively studied, such as LLM-Pruner~\cite{ma2023llm} using Taylor expansion derived from optimal brain damage (OBD)~\cite{lecun1989optimal}, and Sheared-LLaMA~\cite{xia2023sheared}, employing parameterized masks. In contrast, gradient-free methods like TransAct~\cite{shen2024pruning}, typically use activation magnitude for co-designing pruning metrics and architectures, enhancing general capabilities but reducing internal diversity. FLAP~\cite{an2024fluctuation} emphasizes domain-specific diversity, as discussed in Section~\ref{sec:domain-affinity}.

\subsection{Sparse Acceleration}

Beyond static compression, dynamic computation methods like early exiting~\cite{del2023skipdecode} and layer skipping~\cite{raposo2024mixture} accelerate LLM inference by reducing model depth, though they may increase memory demands. To address this, sparsity-aware offloading techniques (e.g., LLM-in-a-flash~\cite{alizadeh2023llm} and PowerInfer~\cite{song2023powerinfer}) exploit sparse activation patterns for efficient computation. Given the successful deployment and acceleration of the Mixtral MoE LLM~\cite{xue2024powerinfer}, DIVE appears highly compatible with these offloading strategies.

\subsection{From Dense to Sparse}

While dense LLMs dominate the landscape, agilely derive an MoE LLM from existing dense LLMs minimize the sunk cost. MoEfication~\cite{zhang2021moefication} introduces expert construction strategies for T5, including random splitting, parameter clustering, and co-activation graph methods. On decoder-only architectures, sparse up-cycling~\cite{komatsuzaki2022sparse} initializes experts using FFN modules, as demonstrated by Skywork-MoE~\cite{wei2024skywork}, which expands a 13B dense LLM into a 146B MoE model with 16 diverse experts. Expert diversity proves crucial, as diverse experts outperform duplicated ones despite higher early-stage training loss. LLaMA-MoE~\cite{zhu2024llama} advocates for a random split approach which divide the dense FFNs into non-overlapping experts.

%% file: body/conclusion.tex
\section{Conclusion}

In this work, we propose DIVE, a diversity-enhanced MoE reconstruction method which significantly reduces the training cost. The three-phase framework of DIVE consists of domain affinity mining, pruning-based expert reconstruction, and efficient retraining, enabling a seamless and effective reconstruction from dense to MoE LLMs. Experiments on language modeling and downstream tasks validate the efficacy of our approach with perplexity reduction and improvements on task accuracy. The analysis of diversity validates the correlation between pruning and MoE expert diversity in enhancing overall performance, while ablation studies highlight the robustness of our method.

%% file: body/after_main_paper.tex
\section*{Limitations}

Although DIVE demonstrates generalization across different models and sizes, the models used in experiments are no larger than 7B due to resource constraints. Future efforts could focus on scaling up model sizes and exploring broader domains to further validate the effectiveness of DIVE, as splitting-based methods and up-cycling have proven effective in larger models.

\section*{Ethics Statement}

Our work focuses on efficiently reconstructing dense LLMs into MoE architectures while enhancing the diversity of the reconstructed models. The datasets used in our study are publicly available and widely recognized in the research community, sourced from open-source repositories. No proprietary or sensitive data were used during either training or evaluation. Moreover, our methods are purely algorithmic and do not promote or amplify harmful biases by design.

\section*{Acknowledgement}

We thank the anonymous reviewers for their insightful feedback, which greatly improves our paper. This work is supported by the National Natural Science Foundation of China (No. 62472419, 62472420).

%% file: body/appendix.tex
\section{Baseline Details}
\label{sec:appendix-a}

\subsection{Expert Reconstruction of LLaMA-MoE}
\label{sec:appendix-a-1}

LLaMA-MoE~\cite{zhu2024llama} presents two groups of expert reconstruction methods: Neuron-Independent and Neuron-Sharing, with each category containing two distinct methods. We initially consider selecting the best method from each group, specifically Independent$_\text{Random}$ and Sharing$_\text{Inter}$, based on their reported performance. However, due to the restriction of Sharing$_\text{Inter}$, which requires setting aside neurons shared by most experts as independent residual blocks, it is not suitable for reconstructing 1/8 MoE models. So we adapt Independent$_\text{Random}$, which is originally supports reconstructing models with a fixed total number of parameters. We make revision to allow it to be adaptive to varying FFN sizes and to enable a fair comparison.

We redefine expert reconstruction as the task of partitioning intermediate neurons into equal-sized subsets. Given a universal set $U$ that contains the indices of all intermediate neurons $\left\{1,2,\dots,d_h\right\}$, we randomly divide $U$ into equal sized $n$ subsets $S_1, S_2, \dots, S_n$, potentially constructing experts of varying sizes as:
\begin{equation}
    \begin{aligned}
        \bigcup_{i=1}^{n} S_i = U. \\
    \end{aligned}
\label{eq:weight}
\end{equation}

In this context, we omit the original condition $\quad \bigcap_{i=1}^{n} S_i = \varnothing$ from LLaMA-MoE.

\input{tables/flap}

\subsection{Discussion on FLAP}
\label{sec:appendix-a-2}

For the structured pruning method in FLAP~\cite{an2024fluctuation}, given a calibration set $C$ and $N$ samples, we first compute the baseline fluctuation of the $j$-th channel in the input feature of layer $\ell$ with the average sample fluctuation variance:
\begin{equation}
    \overline{\mathbf{X}}^{\ell}_{:,j,:,C} = \frac{1}{NL} \sum_{n=1}^{N} \sum_{k=1}^{L} \mathbf{X}^{\ell}_{n,j,k,C}.
\end{equation}

Then, we calculate the fluctuation variance for each input feature in layer $\ell$, which gives the importance score $\mathbf{S}^{\ell}_{:,j,C}$ for the weight group $\mathbf{W}^\ell_{:,j}$:
\begin{equation}
    \mathbf{S}^{\ell}_{:,j,C} = \frac{1}{N-1} \sum_{n=1}^N (\mathbf{X}^{\ell}_{n,j,:,C} - \overline{\mathbf{X}}^{\ell}_{:,j,:,C})^2  \cdot ||\mathbf{W}^{\ell}_{:,j}||^2_2.
    \label{eq:fluc}
\end{equation}

The pruning mask $\mathbf{M}^\ell_{C}$ is determined using these importance scores, by retaining the channels with the highest scores. The output of each pruned layer can be formulated as:
\begin{equation}
    \mathbf{W}^{\ell}_{C} \mathbf{X}^{\ell}_{C} \approx (\mathbf{M}^{\ell}_{C} \odot \mathbf{W}^{\ell}) \mathbf{X}^{\ell}_{C} + \mathbf{B}^{\ell}_{0,C},
\end{equation}
with $\mathbf{B}^{\ell}_{0,C} = \mathbf{W}^{\ell} ((1 - \mathbf{M}^{\ell}_{C}) \odot \overline{\mathbf{X}}^{\ell}_{C})$ representing the bias of the relevant linear layer.

For all pruning processes, we set the calibration size to 1024 and the sample length to 256. To assess the impact of the calibration set, we conduct an experiment comparing language modeling and downstream task performance using FLAP pruning on C4 and a mixed set of 24 datasets, alongside our DIVE 1/8 with only routers trained.

As shown in Table~\ref{tab:flap}, FLAP models pruned on different calibration sets show inconsistent performance across tasks. Specifically, FLAP (C4) lags behind FLAP (Mixed) by up to 9.90 in perplexity of LAMBADA and WikiText2 (sequence length 1024), but outperforms it by 1.39\% on the average results of downstream tasks. In contrast, DIVE 1/8 achieves the best performance in both language modeling (perplexity) and downstream task (average accuracy), benefiting from the diversity of experts. This highlights the sensitivity of existing pruning methods and underscores the effectiveness of our approach.

\subsection{Discussion on LLaMA-MoE}

For Neuron-Sharing methods of LLaMA-MoE, which rely similarly on a pruning-based approach to construct experts and clusters data via sentence embeddings, they prune the model based solely on clustering results from the same dataset, without acknowledging the impact of different datasets. This limitation mirrors the issue we discuss with FLAP above, which can be summarized as a lack of diversity.

In contrast, our method focuses on the insight of the diversity of pruning outcomes, achieving better performance on various tasks. And we explore to reconstruct 1/8 MoE models and adjust the retraining phase to resolve the sparse gradients, especially for the top-$1$ routers.

\section{Retraining Details}
\label{sec:appendix-b}

\subsection{Hyper-parameters}
\label{sec:appendix-b-1}

\input{tables/parameters}

The training parameters for the sparse training stage are presented in Table~\ref{tab:parameters}. Our experiments are conducted on 4 A800 (80G) GPUs with a global batch size of 512, utilizing BFloat16 precision. We set the warmup ratio to 3\% and employ a cosine learning rate scheduler, decaying the learning rate from 1e-4 to 1e-5. Each model is retrained on 5B tokens. Our implementation is based on Huggingface Transformers~\cite{wolf2020transformers} and DeepSpeed~\cite{rasley2020deepspeed}.

For the dense training stage of MoE reconstruction methods, we specifically set a constant learning rate of 1e-4, with a global batch size of 256. We adjust each temperature coefficient ($t$) based on their validation loss, as shown in Table~\ref{tab:temp}. For all MoE models, we do not adopt the re-scaling factor in LLaMA-MoE.

\input{tables/temp}

\begin{figure*}[htbp]
    \centering
    \includegraphics[width=0.95\textwidth]{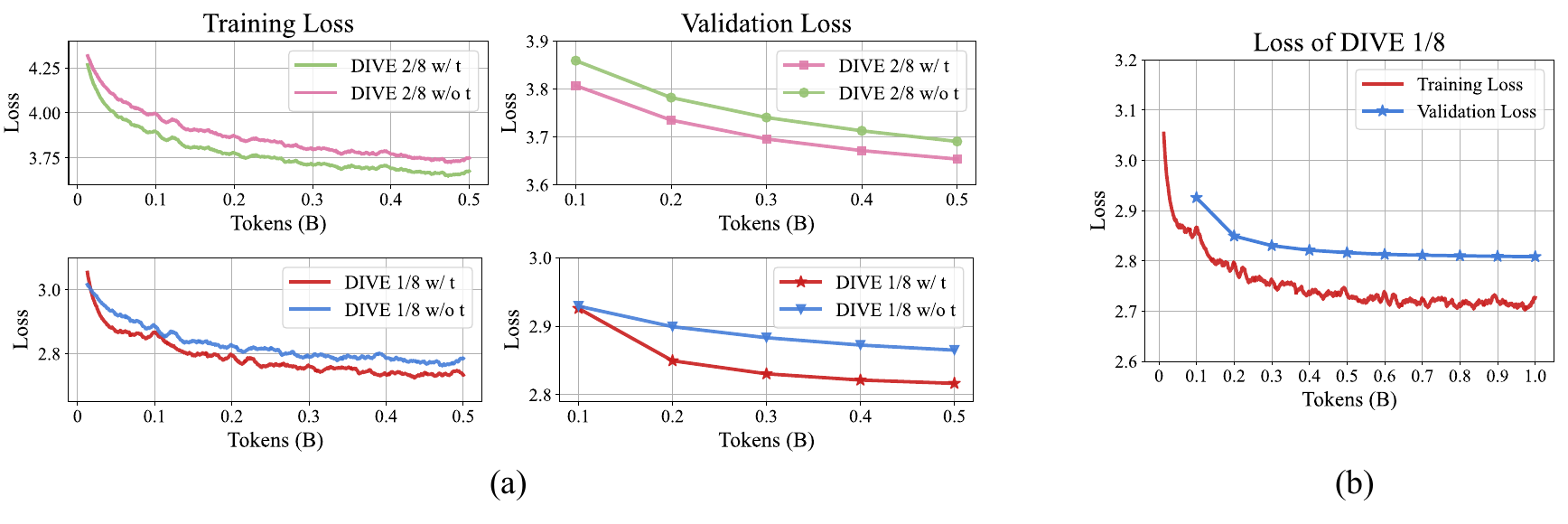}
    \caption{Training and validation loss for (a) DIVE 1/8 and DIVE 2/8 with and without temperature coefficient ($t$), trained on 0 to 0.5B tokens, and (b) DIVE 1/8 with $t$, trained on 0 to 1B tokens.}
\label{fig:router_training}
\end{figure*}

\input{tables/llama_moe}

\subsection{Clarification of Two-Stage Retraining}
\label{sec:appendix-b-2}

As our models are reconstructed based on pruning results from pre-trained LLMs and the abilities of experts are partially retained, it is straightforward to separate the retraining process into two stages to better optimize the different reconstructed modules (routers and experts).

\begin{enumerate}
    \item \textbf{Dense training (for routers)}: Due to the retained abilities of reconstructed experts, the issues of diversity and generalization of pruning can be addressed by focusing sorely on dense router training, as discussed in Appendix~\ref{sec:appendix-a-2}. Activating all experts can deal with the sparse gradients and non-differentiability of the top-$k$ operation in routers.
    \item \textbf{Sparse training (for experts)}: To further enhance the model’s capabilities, we focus on training the experts and the associated normalization modules. With reasonable routing allocations already established, only the appropriate expert need to be activated during each forward and backward pass.
\end{enumerate}

Based on the considerations above, we design a two-stage retraining approach of dense-then-sparse training, which is more suitable for our framework than an end-to-end training process.

\subsection{Temperature Coefficient}
\label{sec:appendix-b-3}

In this section, we explain the rationale behind our temperature coefficient settings. The original routing logits of a sparse MoE are defined as:
\begin{equation}
    w_i(x) = \mathrm{Softmax}(\mathrm{TopK}(z(x)))_i,
\end{equation}
which leads to a gap between the propagation during dense training and the sparse inference. To address this, we introduce a temperature coefficient $t$ to sharpen the logits and approximate the top-$k$ operation, ensuring alignment between the routing behaviors. The router output is modified as:
\begin{equation}
    w_i(x) = \mathrm{Softmax}\left(\frac{z(x)}{t}\right)_i.
   \label{eq:top_k}
\end{equation}

Figure~\ref{fig:router_training} (a) shows the training and validation loss throughout dense training of routers, from 0 to 0.5B tokens. For both DIVE 1/8 and DIVE 2/8, the models with the temperature coefficient consistently outperform those without it (validation loss of 2.82 vs 2.87, 3.65 vs 3.69), which aligns with our expectations. Furthermore, as shown in Figure~\ref{fig:router_training} (b), the models converge around 0.5B tokens, which is the point we use to standardize the dense training stage.

\section{Further Retraining Details}
\label{sec:appendix-c}

\subsection{Hyper-parameters}

To further validate our approach, we retrain the DIVE and LLaMA-MoE models on an expanded training dataset. The experiments are conducted using 4 A800 (80GB) GPUs with a global batch size of 256 and a fixed learning rate of 1e-5. Each model is further trained on an additional 10B tokens, resulting in a total of 15B tokens including the initial retraining. Detailed parameters are provided in Table~\ref{tab:parameters_further}; unless otherwise specified, the settings are consistent with those in Appendix~\ref{sec:appendix-b-1}.

\input{tables/parameters_further}

\begin{figure*}[htbp]
    \centering
    \includegraphics[width=1\linewidth]{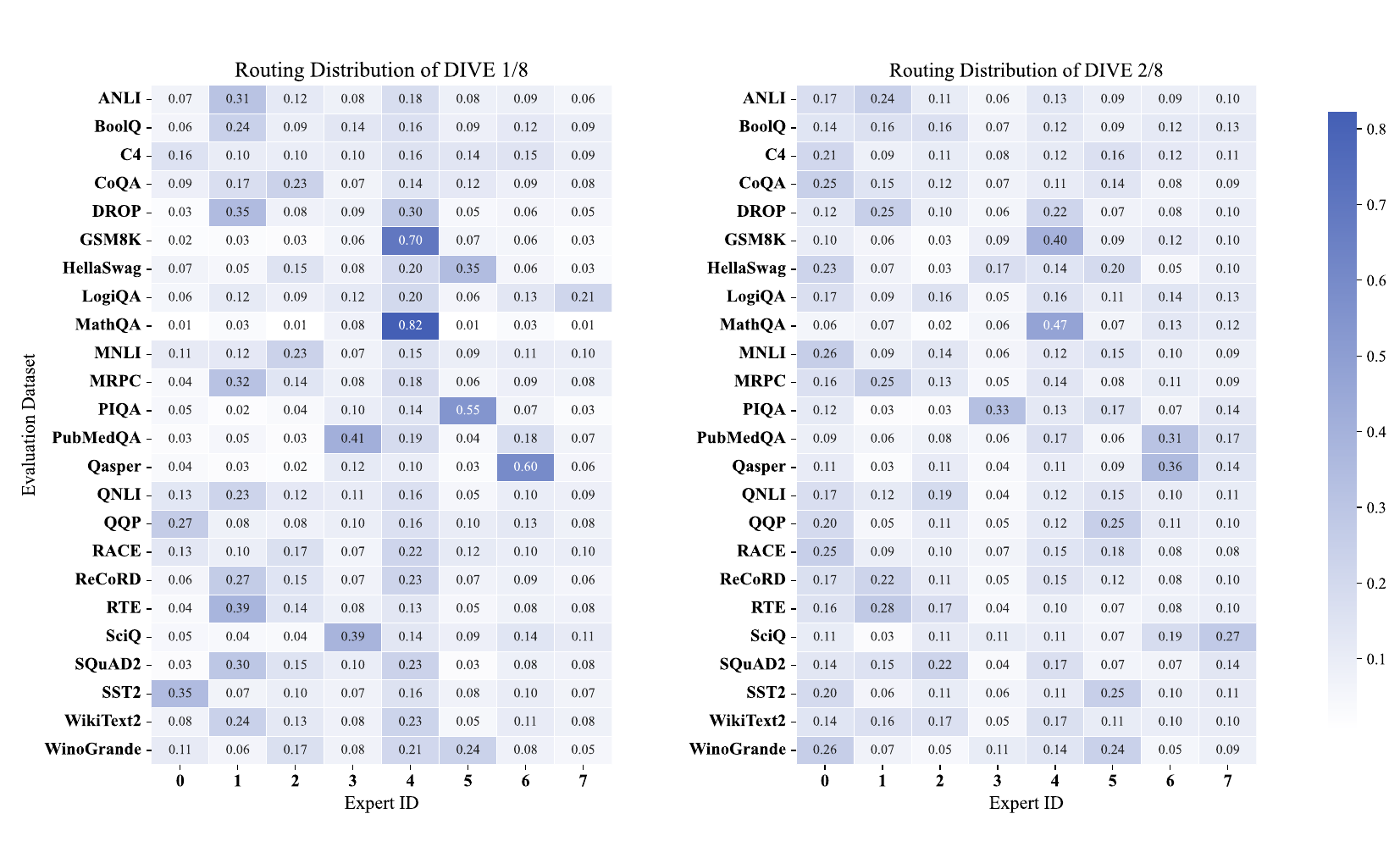}
    \caption{Full routing distribution of DIVE 1/8 (left) and DIVE 2/8 (right), shown as activation ratios of the corresponding experts across all layers. The indexes of experts correspond to the domain clustering results in Appendix~\ref{sec:appendix-d-2}.}
\label{fig:routing_full}
\end{figure*}

\subsection{Results}

Table~\ref{tab:llama_moe} reports the performance of DIVE and LLaMA-MoE models after retraining on the extended token budget. The first two columns, marked with downward arrows, report perplexity on language modeling benchmarks, while the remaining columns present accuracy results on a suite of downstream tasks.

Among the four models evaluated, DIVE 1/8 achieves the best overall performance, with the lowest perplexity scores on both WikiText2 and LAMBADA, and the highest average accuracy across downstream tasks. In particular, it reduces the perplexity of the LLaMA-MoE models by at least 5.37 on WikiText2 and 49.81 on LAMBADA, and improves the best average downstream accuracy of LLaMA-MoE by 2.75\%. It shows strong performance on SciQ, PIQA, ARC-e, ARC-c, HellaSwag, and OBQA. DIVE 2/8 achieves the second-best overall results, with notable improvements in WinoGrande, LogiQA, and MMLU. These results demonstrate that our approach consistently outperforms LLaMA-MoE when retrained with more data, highlighting the robustness of our method.

\input{tables/affinity_full}

\section{Domain Clustering Details}
\label{sec:appendix-d}

\subsection{Domain Affinity Mining Datasets}
\label{sec:appendix-d-1}

In domain affinity mining, we conduct a large-scale empirical study on the relationship between $24$ calibration datasets used for pruning and the performance of the pruned models. Our dataset selection is inspired by SPoT~\cite{vu2021spot}, a transfer learning approach using a learned prompt in source tasks and then initializing the prompt for a target task, which gains superior performance in a variety of downstream tasks, demonstrating its generalization. Considering generalization and an exploration of specialization, we use the source task groups in SPoT and extend them by specialized tasks, using their training sets as calibration datasets and test sets as the evaluation datasets.

These datasets include language modeling (C4~\cite{raffel2020exploring}, WikiText2~\cite{merity2016pointer}), natural language inference (ANLI~\cite{nie2019adversarial}, MNLI~\cite{williams2017broad}, QNLI~\cite{zhang2015character}, RTE~\cite{wang2018glue}), question answering (SQuAD2~\cite{rajpurkar-etal-2018-know}, BoolQ~\cite{clark-etal-2019-boolq}, PIQA~\cite{bisk2020piqa}, DROP~\cite{dua2019drop}, CoQA~\cite{reddy2019coqa}, LogiQA~\cite{liu2020logiqa}, ReCoRD~\cite{zhang2018record}), math problems (GSM8K~\cite{cobbe2021training}, MathQA~\cite{amini-etal-2019-mathqa}), specialized knowledge question answering (SciQ~\cite{SciQ}, Qasper~\cite{dasigi2021dataset}, PubMedQA~\cite{jin2019pubmedqa}, RACE~\cite{lai2017race}), commonsense reasoning (WinoGrande~\cite{WINOGRANDE}, HellaSwag~\cite{zellers2019hellaswag}), paraphrase detection (QQP~\cite{wang2018glue}, MRPC~\cite{dolan2005automatically}), sentiment analysis (SST-2~\cite{socher2013recursive}).

\subsection{Domain Clustering Results}
\label{sec:appendix-d-2}

We apply our domain-specific data selection to various dense LLM backbones with different activation functions, including LLaMA2-7B, TinyLlama-1.1B, OPT-6.7B and Qwen2.5-7B, resulting in 8 hierarchical clusters on each LLM with FFN pruning ratios of 50\% and 75\%. These clusters are used to define the domains of experts in the reconstruction phase. The clustering results are presented in Table~\ref{tab:clusters}. Each pruning ratio is inversely related to the size of a single FFN, where $\text{FFN Size} = 1 - \text{FFN Pruning Ratio}$.

\section{Diversity Analysis Details}
\label{sec:appendix-e}

\subsection{Routing Distribution}

As described in Section~\ref{sec:exp-diversity}, we analyze routing distributions across 24 pruning datasets to assess expert diversity. The complete heatmaps are shown in Figure~\ref{fig:routing_full}, where each value indicates the normalized activation ratio of tokens routed to a specific expert. Expert indices are aligned with domain clusters. These results demonstrate the effectiveness of our retraining setup and the successful realization of diverse expert routing.

\subsection{Case Study}

We do a case study on routing distribution to further analyze the diversity, specifically on DIVE 1/8. In Table~\ref{tab:case_study}, we select data from the test set of four datasets: C4, MathQA, GSM8K and PubMedQA, and analyze the routing results. The original sentences and expertised ones, which are tokenized and routed, are shown. The corresponding expert is determined by the expert with the highest frequency of the current token across all 22 layers of TinyLlama-1.1B.

As expected, the tokens in the C4 dataset is routed relatively evenly across experts, while the data in MathQA, GSM8K and PubMedQA tends to be more concentrated in their corresponding domain-specific experts. Specifically, many technical terms, such as "Neuronal" and all numerical expressions, are routed to the appropriate experts (Expert 3 and Expert 4). This confirms the success of our diversity approach and routing distribution.

\section{Ablation Details}
\label{sec:appendix-f}

In Section~\ref{sec:exp-domain}, to intuitively demonstrate the contribution of domain affinity mining (D.A.M) to DIVE, downstream tasks are categorized into in-distribution (ID) and out-of-distribution (OOD) groups, based on whether they belong to the 24 datasets used for pruning. Specifically, ID tasks include SciQ, PIQA, WinoGrande, MathQA, HellaSwag, LogiQA, and BoolQ, while OOD tasks include ARC-e, ARC-c, OBQA, and MMLU.

Besides the ablation study on DIVE 2/8, we also conduct one on DIVE 1/8 with and without D.A.M. The full results are shown in Table~\ref{tab:affinity-full}. These results indicate that incorporating D.A.M into DIVE 1/8 reduces perplexity on WikiText2 and LAMBADA by 0.35 and 3.99, respectively. Furthermore, downstream task performance of DIVE 1/8 consistently improves with D.A.M, highlighting its significant role and impact on overall model effectiveness.

\input{tables/clusters}

\input{tables/case}

%% file: tables/flap.tex
\begin{table*}[t]
    \centering
    \resizebox{\textwidth}{!}{%
    \begin{tabular}{lcccccccc}
        \toprule
        \textbf{Method} & \textbf{WikiText2 $\downarrow$} &\textbf{LAMBADA $\downarrow$} & \textbf{SciQ} & \textbf{PIQA} & \textbf{WinoGrande} & \textbf{ARC-e} & \textbf{ARC-c (25)} \\
        \midrule
        FLAP (C4) & 26.32 & 70.27 & \underline{84.30} & 60.94 & \underline{53.67} & 40.40 & 23.12 \\
        FLAP (Mixed) & \underline{22.67} & \underline{60.37} & \textbf{85.20} & \underline{62.40} & \textbf{53.99} & \underline{44.28} & \underline{23.21} \\
        DIVE 1/8 & \textbf{17.30} & \textbf{42.55} & 83.60 & \textbf{65.72} & 52.96 & \textbf{48.57} & \textbf{25.09} \\
        \midrule
        \textbf{Method} & \textbf{MathQA} & \textbf{HellaSwag (10)} & \textbf{LogiQA} & \textbf{BoolQ} & \textbf{OBQA} & \textbf{MMLU (5)} & \textbf{Average} \\
        \midrule
        FLAP (C4) & 22.61 & \underline{33.66} & 20.28 & \textbf{62.29} & \underline{17.00} & \underline{23.01} & \underline{40.12} \\
        FLAP (Mixed) & \textbf{23.62} & 32.83 & \underline{20.89} & 39.91 & 16.80 & 22.92 & 38.73 \\
        DIVE 1/8 & \underline{23.15} & \textbf{35.71} & \textbf{22.43} & \underline{51.50} & \textbf{17.60} & \textbf{26.16} & \textbf{41.14} \\
        \bottomrule
    \end{tabular}
    }
    \caption{Performance of FLAP pruned on C4 and mixed data, and DIVE 1/8 reconstructed without sparse training. Tasks marked with downward arrows (WikiText2 and LAMBADA) are evaluated by perplexity, which is inversely correlated with performance. Others are downstream tasks measured by accuracy, with averages shown in the last column. The best results are in \textbf{bold}, and the second-best results are \underline{underlined}.}
    \label{tab:flap}
\end{table*}

%% file: tables/parameters.tex
\begin{table}[htbp]
    \centering
    \begin{tabularx}{0.4\textwidth}{Xc}
        \toprule
        \textbf{Training Parameter} & \textbf{Value} \\
        \midrule
        \# GPUs & 4 \\
        Sequence length & 1024 \\
        Data type & Bfloat16 \\
        Learning rate & 1e-4 \\
        Learning rate scheduler & Cosine \\
        Cosine minimum ratio & 0.1 \\
        Warmup ratio & 0.03 \\
        Optimizer & AdamW \\
        Global batch size & 512 \\
        LoRA rank & 8 \\
        LoRA alpha & 16 \\
        LoRA dropout & 0.1 \\
        DeepSpeed & Zero-2 \\
        \bottomrule
    \end{tabularx}
    \caption{Training parameters used in the experiments.}
    \label{tab:parameters}
\end{table}

%% file: tables/temp.tex
\begin{table}[htbp]
    \centering
    \begin{tabularx}{0.32\textwidth}{Xc}
        \toprule
        \textbf{Method} & \textbf{$t$} \\
        \midrule
        DIVE 1-8 & 0.05 \\
        DIVE 2-8 & 0.5 \\
        LLaMA-MoE 1-8 & 0.01 \\
        LLaMA-MoE 2-8 & 0.1 \\
        \bottomrule
    \end{tabularx}
    \caption{Temperature coefficient($t$) for MoE reconstruction methods used in dense training.}
    \label{tab:temp}
\end{table}

%% file: tables/llama_moe.tex
\begin{table*}[t]
    \centering
    \resizebox{\textwidth}{!}{%
    \begin{tabular}{lcccccccc}
        \toprule
        \textbf{Method} & \textbf{WikiText2 $\downarrow$} &\textbf{LAMBADA $\downarrow$} & \textbf{SciQ} & \textbf{PIQA} & \textbf{WinoGrande} & \textbf{ARC-e} & \textbf{ARC-c (25)} \\
        \midrule
        LLaMA-MoE 1/8 & 18.60 & 73.96 & 77.40 & 61.15 & 52.81 & 43.56 & 22.10 \\
        DIVE 1/8 & \textbf{13.23} & \textbf{24.15} & \textbf{83.00} & \textbf{68.01} & \underline{54.70} & \textbf{52.53} & \textbf{24.40} \\
        \hdashline
        LLaMA-MoE 2/8 & 19.45 & 96.80 & 77.90 & 62.84 & 53.51 & 42.89 & 22.01 \\
        DIVE 2/8 & \underline{15.87} & \underline{47.68} & \underline{81.50} & \underline{64.96} & \textbf{54.78} & \underline{47.10} & \underline{23.12} \\
        \midrule
        \textbf{Method} & \textbf{MathQA} & \textbf{HellaSwag (10)} & \textbf{LogiQA} & \textbf{BoolQ} & \textbf{OBQA} & \textbf{MMLU (5)} & \textbf{Average} \\
        \midrule
        LLaMA-MoE 1/8 & 21.07 & \underline{33.73} & 20.58 & \underline{61.71} & 16.60 & 25.47 & 39.65 \\
        DIVE 1/8 & \underline{22.55} & \textbf{37.17} & 19.35 & 59.88 & \textbf{20.20} & \underline{25.88} & \textbf{42.52} \\
        \hdashline
        LLaMA-MoE 2/8 & \textbf{23.08} & 32.23 & \underline{21.51} & \textbf{61.77} & 14.60 & 25.14 & 39.77 \\
        DIVE 2/8 & 22.58 & 33.27 & \textbf{21.97} & 58.47 & \underline{18.20} & \textbf{26.33} & \underline{41.12} \\
        \bottomrule
    \end{tabular}
    }
    \caption{Performance of LLaMA-MoE and DIVE models further retrained on 15B tokens. Tasks marked with downward arrows (WikiText2 and LAMBADA) are evaluated by perplexity with a sequence length of 1024, and others are downstream tasks measured by accuracy. The last column shows the average accuracy of all downstream tasks. The best results are in \textbf{bold}, and the second-best results are \underline{underlined}.}
    \label{tab:llama_moe}
\end{table*}

%% file: tables/parameters_further.tex
\begin{table}[htbp]
    \centering
    \begin{tabularx}{0.4\textwidth}{Xc}
        \toprule
        \textbf{Training Parameter} & \textbf{Value} \\
        \midrule
        \# GPUs & 4 \\
        Learning rate & 1e-5 \\
        Learning rate scheduler & Constant \\
        Global batch size & 256 \\
        \bottomrule
    \end{tabularx}
    \caption{Training parameters for the further retraining experiments (an additional 10B-token training phase, totaling 15B tokens), applied to DIVE and LLaMA-MoE models.}
    \label{tab:parameters_further}
\end{table}

%% file: tables/affinity_full.tex
\begin{table*}[t]
    \centering
    \resizebox{\textwidth}{!}{%
    \begin{tabular}{lcccccccc}
        \toprule
        \textbf{Method} & \textbf{WikiText2 $\downarrow$} &\textbf{LAMBADA $\downarrow$} & \textbf{SciQ} & \textbf{PIQA} & \textbf{WinoGrande} & \textbf{MathQA} & \textbf{HellaSwag} \\
        \midrule
        DIVE 1/8 & \textbf{13.90} & \textbf{27.09} & \textbf{82.90} & \textbf{67.19} & \textbf{55.01} & \textbf{22.55} & \textbf{36.97} \\
        - D.A.M & 14.25 & 31.08 & 81.60 & 65.40 & 54.62 & 22.35 & 35.84 \\
        \hdashline
        DIVE 2/8 & \textbf{18.09} & \textbf{63.45} & \textbf{81.50} & \textbf{63.98} & \textbf{52.96} & \textbf{22.65} & \textbf{32.26} \\
        - D.A.M & 20.02 & 76.03 & 79.90 & 60.72 & 52.33 & 22.48 & 30.40 \\
        \midrule
        \textbf{Method} & \textbf{LogiQA} & \textbf{BoolQ} & \textbf{ARC-e} & \textbf{ARC-c (25)} & \textbf{OBQA} & \textbf{MMLU (5)} & \textbf{Average} \\
        \midrule
        DIVE 1/8 & 18.59 & 59.54 & \textbf{51.56} & 25.94 & 18.80 & \textbf{25.10} & \textbf{42.20} \\
        - D.A.M & \textbf{20.74} & \textbf{61.31} & 50.67 & \textbf{26.11} & \textbf{20.60} & 23.86 & 42.10 \\
        \hdashline
        DIVE 2/8 & 21.66 & 60.18 & \textbf{44.87} & \textbf{22.18} & 16.60 & \textbf{27.03} & \textbf{40.53} \\
        - D.A.M & \textbf{21.81} & \textbf{61.31} & 40.74 & 21.67 & \textbf{17.20} & 24.98 & 39.47 \\
        \bottomrule
    \end{tabular}
    }
    \caption{Performance of DIVE with and without D.A.M, retrained on 1B tokens. Tasks marked with downward arrows (WikiText2 and LAMBADA) are evaluated by perplexity with a sequence length of 1024, and others are downstream tasks measured by accuracy. The last column shows the average accuracy of all downstream tasks. The best results are in \textbf{bold}.}
    \label{tab:affinity-full}
\end{table*}

%% file: tables/clusters.tex
\begin{table*}[htbp]
    \centering
    \resizebox{\textwidth}{!}{
    \begin{tabular}{lccl}
        \toprule
        \textbf{Dense Model} & \textbf{FFN Pruning Ratio} & \textbf{Cluster Index} & \textbf{Calibration Datasets} \\
        \midrule
        \multirow{8}{*}{LLaMA2-7B} & \multirow{8}{*}{50\%} & 0 & C4, MNLI, ANLI, RACE, CoQA \\
            & & 1 & QNLI, SQuAD2, BoolQ \\
            & & 2 & WikiText2, RTE, DROP, ReCoRD, MRPC \\
            & & 3 & SciQ, Qasper, PubMedQA \\
            & & 4 & GSM8K, MathQA \\
            & & 5 & PIQA, HellaSwag \\
            & & 6 & WinoGrande, SST-2 \\
            & & 7 & LogiQA, QQP \\
        \midrule
        \multirow{8}{*}{LLaMA2-7B} & \multirow{8}{*}{75\%} & 0 & C4, MNLI, ANLI, CoQA \\
            & & 1 & QNLI, SQuAD2, BoolQ \\
            & & 2 & WikiText2, RTE, DROP, ReCoRD, MRPC \\
            & & 3 & SciQ, Qasper, PubMedQA \\
            & & 4 & GSM8K, MathQA \\
            & & 5 & PIQA, WinoGrande, HellaSwag \\
            & & 6 & RACE, QQP, SST-2 \\
            & & 7 & LogiQA \\
        \midrule
        \multirow{8}{*}{TinyLlama-1.1B} & \multirow{8}{*}{50\%} & 0 & C4, RACE, QQP, SST-2 \\
            & & 1 & WikiText2, QNLI, RTE, SQuAD2, BoolQ, DROP, ReCoRD, MRPC \\
            & & 2 & MNLI, ANLI, CoQA \\
            & & 3 & SciQ, PubMedQA \\
            & & 4 & GSM8K, MathQA \\
            & & 5 & PIQA, WinoGrande, HellaSwag \\
            & & 6 & Qasper \\
            & & 7 & LogiQA \\
        \midrule
        \multirow{8}{*}{TinyLlama-1.1B} & \multirow{8}{*}{75\%} & 0 & C4, MNLI, CoQA, RACE \\
            & & 1 & WikiText2, RTE, ANLI, BoolQ, DROP, ReCoRD, MRPC \\
            & & 2 & QNLI, SQuAD2, LogiQA \\
            & & 3 & PIQA \\
            & & 4 & GSM8K, MathQA \\
            & & 5 & WinoGrande, HellaSwag, QQP, SST2 \\
            & & 6 & Qasper, PubMedQA \\
            & & 7 & SciQ \\
        \midrule
        \multirow{8}{*}{OPT-6.7B} & \multirow{8}{*}{50\%} & 0 & C4, LogiQA \\
            & & 1 & WikiText2, QNLI, SQuAD2, BoolQ \\
            & & 2 & MNLI, ANLI, CoQA \\
            & & 3 & RTE, DROP, ReCoRD, MRPC \\
            & & 4 & GSM8K, MathQA, HellaSwag, PIQA \\
            & & 5 & RACE, WinoGrande, SST2 \\
            & & 6 & QQP \\
            & & 7 & SciQ, Qasper, PubMedQA \\
        \midrule
        \multirow{8}{*}{OPT-6.7B} & \multirow{8}{*}{75\%} & 0 & C4, QNLI \\
            & & 1 & WikiText2, SQuAD2, BoolQ \\
            & & 2 & ANLI, CoQA \\
            & & 3 & RTE, DROP, ReCoRD, MRPC \\
            & & 4 & GSM8K, MathQA, HellaSwag, QQP, PIQA \\
            & & 5 & WinoGrande, SST2 \\
            & & 6 & MNLI, RACE \\
            & & 7 & LogiQA, SciQ, Qasper, PubMedQA \\
        \midrule
        \multirow{8}{*}{Qwen2.5-7B} & \multirow{8}{*}{50\%} & 0 & C4, RACE, SST2 \\
            & & 1 & WikiText2, QNLI, SQuAD2, BoolQ, DROP \\
            & & 2 & MNLI, RTE, ANLI, CoQA, ReCoRD, MRPC \\
            & & 3 & PIQA, HellaSwag \\
            & & 4 & GSM8K, MathQA \\
            & & 5 & WinoGrande \\
            & & 6 & SciQ, Qasper, PubMedQA \\
            & & 7 & LogiQA, QQP \\
        \bottomrule
    \end{tabular}
    }
    \caption{Clustering results of LLaMA2-7B, TinyLlama-1.1B, OPT-6.7B and Qwen2.5-7B for FFN pruning ratios of $50\%$ and $75\%$.}
    \label{tab:clusters}
\end{table*}

%% file: tables/case.tex
\begin{table*}[htbp]
    \centering
    \resizebox{\textwidth}{!}{
    \begin{tabular}{p{0.28\textwidth} p{0.72\textwidth}}
        \toprule
        \textbf{Data Source} & \textbf{Sentence} \\
        \midrule
        \textbf{C4 (Original)} & 
        Who am I? What's this all About? You could say I grew up traveling, when I was young I lived in Lagos with my family. Over the years I spent there we traveled to some of the countries you see highlighted in green. \\
        \hdashline
        \textbf{C4 (Expertised)} & 
            \textcolor{red}{<s>} \textcolor{yellow}{\_Who} \textcolor{yellow}{\_am} \textcolor{purple}{\_I} \textcolor{purple}{?} \textcolor{purple}{\_What} \textcolor{yellow}{'} \textcolor{purple}{s} \textcolor{purple}{\_this} \textcolor{purple}{\_all} \textcolor{purple}{\_About} \textcolor{red}{?} \textcolor{yellow}{\_You} \textcolor{purple}{\_could} \textcolor{yellow}{\_say} \textcolor{yellow}{\_I}  \textcolor{purple}{\_grew} \textcolor{purple}{\_up} \textcolor{purple}{\_travel} \textcolor{yellow}{\_ing} \textcolor{yellow}{","} \textcolor{yellow}{\_when} \textcolor{yellow}{\_I} \textcolor{yellow}{\_was} \textcolor{yellow}{\_young} \textcolor{yellow}{\_I} \textcolor{yellow}{\_lived} \textcolor{yellow}{\_in} \textcolor{green}{\_Lag} \textcolor{green}{os} \textcolor{yellow}{\_with} \textcolor{yellow}{\_my} \textcolor{yellow}{\_family} \textcolor{red}{.} \textcolor{yellow}{\_Over} \textcolor{red}{\_the} \textcolor{green}{\_years} \textcolor{yellow}{\_I} \textcolor{purple}{\_spent} \textcolor{yellow}{\_there} \textcolor{purple}{\_we} \textcolor{magenta}{\_tra} \textcolor{green}{ve} \textcolor{purple}{led} \textcolor{green}{\_to} \textcolor{purple}{\_some} \textcolor{cyan}{\_of} \textcolor{green}{\_the} \textcolor{green}{\_countries} \textcolor{yellow}{\_you} \textcolor{yellow}{\_see} \textcolor{cyan}{\_hightlight} \textcolor{cyan}{ed} \textcolor{yellow}{\_in} \textcolor{magenta}{\_green} \textcolor{red}{.} \\
        \midrule
        \textbf{MathQA (Original)} & 
        a multiple choice test consists of 4 questions, and each question has 5 answer choices. in how many r ways can the test be completed if every question is unanswered? \\
        \hdashline
        \textbf{MathQA (Expertised)} & 
            \textcolor{red}{<s>} \textcolor{red}{\_a} \textcolor{cyan}{\_multiple} \textcolor{cyan}{\_choice} \textcolor{cyan}{\_test} \textcolor{cyan}{\_consists} \textcolor{cyan}{\_of} \textcolor{red}{\_} \textcolor{red}{4} \textcolor{cyan}{\_questions} \textcolor{red}{"\_,"} \textcolor{cyan}{\_and} \textcolor{red}{\_each} \textcolor{cyan}{\_question} \textcolor{cyan}{\_has} \textcolor{red}{\_} \textcolor{red}{5} \textcolor{cyan}{\_answer} \textcolor{cyan}{\_choices} \textcolor{red}{\_.} \textcolor{red}{\_in} \textcolor{purple}{\_how} \textcolor{red}{\_many} \textcolor{orange}{\_r} \textcolor{red}{\_ways} \textcolor{red}{\_can} \textcolor{red}{\_the} \textcolor{cyan}{\_test} \textcolor{cyan}{\_be} \textcolor{cyan}{\_completed} \textcolor{red}{\_if} \textcolor{red}{\_every} \textcolor{cyan}{\_question} \textcolor{cyan}{\_is} \textcolor{cyan}{\_un} \textcolor{cyan}{answer} \textcolor{cyan}{ed} \textcolor{red}{\_?} \\
        \midrule
        \textbf{GSM8K (Original)} & 
        To make the pizza, Kimber half as many teaspoons of salt as the number of cups of flour, meaning she needs 1/2*16 = <<16*1/2=8>>8 teaspoons of salt. The total number of cups of flour and teaspoons of salt she needs is 8+16 = <<8+16=24>>24 She also needs 10 cups of water, which means the total number of cups of water and flour and teaspoons of salt she needs is 24+10 = <<24+10=34>>34 \#\#\#\# 34 \\
        \hdashline
        \textbf{GSM8K (Expertised)} & 
            \textcolor{red}{<s>} \textcolor{red}{\_To} \textcolor{red}{\_make} \textcolor{red}{\_the} \textcolor{magenta}{\_p} \textcolor{magenta}{izza} \textcolor{magenta}{","} \textcolor{green}{\_Kim} \textcolor{green}{ber} \textcolor{magenta}{\_half} \textcolor{purple}{\_as} \textcolor{magenta}{\_many} \textcolor{magenta}{\_te} \textcolor{magenta}{as} \textcolor{magenta}{po} \textcolor{magenta}{ons} \textcolor{magenta}{\_of} \textcolor{magenta}{\_salt} \textcolor{magenta}{\_as} \textcolor{red}{\_the} \textcolor{red}{\_number} \textcolor{red}{\_of} \textcolor{magenta}{\_cu} \textcolor{magenta}{ps} \textcolor{magenta}{\_of} \textcolor{magenta}{\_fl} \textcolor{magenta}{our} \textcolor{magenta}{,} \textcolor{cyan}{\_meaning} \textcolor{magenta}{\_she} \textcolor{red}{\_needs} \textcolor{red}{\_} \textcolor{red}{1} \textcolor{red}{/} \textcolor{red}{2} \textcolor{red}{*} \textcolor{red}{1} \textcolor{red}{6} \textcolor{red}{\_=} \textcolor{red}{\_<<} \textcolor{red}{1} \textcolor{red}{6} \textcolor{red}{*} \textcolor{red}{1} \textcolor{red}{/} \textcolor{red}{2} \textcolor{red}{=} \textcolor{red}{8} \textcolor{red}{>>} \textcolor{red}{8} \textcolor{orange}{\_te} \textcolor{magenta}{as} \textcolor{orange}{po} \textcolor{magenta}{ons} \textcolor{red}{\_of} \textcolor{magenta}{\_salt} \textcolor{red}{.} \textcolor{red}{\_The} \textcolor{cyan}{\_total} \textcolor{red}{\_number} \textcolor{red}{\_of} \textcolor{red}{\_cu} \textcolor{red}{ps} \textcolor{red}{\_of} \textcolor{orange}{\_fl} \textcolor{magenta}{our} \textcolor{red}{\_and} \textcolor{orange}{\_te} \textcolor{magenta}{as} \textcolor{magenta}{po} \textcolor{magenta}{ons} \textcolor{red}{\_of} \textcolor{red}{\_salt} \textcolor{magenta}{\_she} \textcolor{red}{\_needs} \textcolor{red}{\_is} \textcolor{red}{\_} \textcolor{red}{8} \textcolor{red}{+} \textcolor{red}{1} \textcolor{red}{6} \textcolor{red}{\_=} \textcolor{red}{\_<<} \textcolor{red}{8} \textcolor{red}{+} \textcolor{red}{1} \textcolor{red}{6} \textcolor{red}{=} \textcolor{red}{2} \textcolor{red}{4} \textcolor{red}{>>} \textcolor{red}{2} \textcolor{red}{4} \textcolor{red}{\_She} \textcolor{red}{\_also} \textcolor{red}{\_needs} \textcolor{orange}{\_} \textcolor{magenta}{1} \textcolor{magenta}{0} \textcolor{magenta}{\_cu} \textcolor{red}{ps} \textcolor{red}{of} \textcolor{red}{\_water} \textcolor{red}{","} \textcolor{red}{\_which} \textcolor{red}{\_means} \textcolor{red}{\_the} \textcolor{magenta}{\_total} \textcolor{red}{\_number} \textcolor{red}{\_of} \textcolor{magenta}{\_cu} \textcolor{red}{ps} \textcolor{magenta}{\_of} \textcolor{magenta}{\_water} \textcolor{red}{\_and} \textcolor{orange}{\_fl} \textcolor{magenta}{our} \textcolor{magenta}{\_and} \textcolor{red}{\_te} \textcolor{magenta}{as} \textcolor{red}{po} \textcolor{red}{ons} \textcolor{red}{\_of} \textcolor{red}{\_salt} \textcolor{red}{\_she} \textcolor{red}{\_needs} \textcolor{red}{\_is} \textcolor{red}{\_} \textcolor{red}{2} \textcolor{red}{4} \textcolor{red}{+} \textcolor{red}{1} \textcolor{red}{0} \textcolor{red}{\_=} \textcolor{red}{\_<<} \textcolor{red}{2} \textcolor{red}{4} \textcolor{red}{+} \textcolor{red}{1} \textcolor{red}{0} \textcolor{red}{=} \textcolor{red}{3} \textcolor{red}{4} \textcolor{red}{\_\#\#\#\#} \textcolor{red}{\_} \textcolor{red}{3} \textcolor{red}{4} \\
        \midrule
        \textbf{PubMedQA (Original)} & 
        Neuronal signals via the hepatic vagus nerve contribute to the development of steatohepatitis and protection against obesity in HFD fed Pemt(-/-) mice. \\
        \hdashline
        \textbf{PubMedQA (Expertised)} & 
            \textcolor{red}{<s>} \textcolor{orange}{\_Ne} \textcolor{orange}{ur} \textcolor{orange}{onal} \textcolor{orange}{\_signals} \textcolor{orange}{\_via} \textcolor{orange}{\_the} \textcolor{orange}{\_he} \textcolor{orange}{p} \textcolor{orange}{\_atic} \textcolor{orange}{\_v} \textcolor{orange}{ag} \textcolor{orange}{us} \textcolor{orange}{\_n} \textcolor{orange}{erve} \textcolor{cyan}{\_contribute} \textcolor{orange}{\_to} \textcolor{orange}{\_the} \textcolor{orange}{\_development} \textcolor{orange}{\_of} \textcolor{orange}{\_ste} \textcolor{orange}{ato} \textcolor{orange}{he} \textcolor{orange}{pat} \textcolor{orange}{itis} \textcolor{orange}{\_and} \textcolor{orange}{\_protection} \textcolor{blue}{\_against} \textcolor{orange}{\_ob} \textcolor{orange}{es} \textcolor{orange}{ity} \textcolor{orange}{\_in} \textcolor{orange}{\_H} \textcolor{orange}{FD} \textcolor{orange}{\_fed} \textcolor{orange}{\_P} \textcolor{orange}{em} \textcolor{orange}{t} \textcolor{orange}{(-} \textcolor{red}{/} \textcolor{orange}{-)} \textcolor{orange}{\_m} \textcolor{orange}{ice} \textcolor{red}{.} \\
        \bottomrule
    \end{tabular}
    }
    \caption{Case study on DIVE 1/8. (Original) refers to the original data, while the (Expertised) ones denote the routing results to experts of tokenized sentences and are marked by colors. \textcolor{purple}{Purple}, \textcolor{green}{green}, \textcolor{yellow}{yellow}, \textcolor{orange}{orange}, \textcolor{red}{red}, \textcolor{magenta}{magenta}, \textcolor{cyan}{cyan} and \textcolor{blue}{blue} respectively denote the Expert IDs of 0-7.}
    \label{tab:case_study}
\end{table*}